\documentclass[letterpaper]{article} 
\usepackage{aaai24}  
\usepackage{times}  
\usepackage{helvet}  
\usepackage{courier}  
\usepackage[hyphens]{url}  
\usepackage{graphicx} 
\usepackage{natbib}  
\usepackage{caption} 
\usepackage{algorithm}
\usepackage{algorithmic}
\usepackage[dvipsnames]{xcolor}
\usepackage{subcaption}

\usepackage{sourcecodepro}
\usepackage{anyfontsize}
\usepackage{booktabs}
\usepackage{pifont}
\usepackage{multirow}
\usepackage{array}
\usepackage{tikz}
\usepackage{pgfplots}
\usetikzlibrary{patterns}

\usepackage{newfloat}
\usepackage{listings}
\title{Relational Programming with Foundation Models}
\author{
	Ziyang Li,
	Jiani Huang,
	Jason Liu,
	Felix Zhu,
	Eric Zhao,
	William Dodds,
	Neelay Velingker, \\
	Rajeev Alur,
	Mayur Naik
}
\affiliations{

    University of Pennsylvania
    
    liby99@seas.upenn.edu, 
    jianih@seas.upenn.edu, 
    jasonhl@seas.upenn.edu,
    zhufelix@seas.upenn.edu,
    zhaoer@seas.upenn.edu,
    wdodds@sas.upenn.edu,
    neelay@seas.upenn.edu,
    alur@seas.upenn.edu,
    mhnaik@seas.upenn.edu
}

\newcommand\todo[1]{\textcolor{red}{#1}}

\newcommand{\cmark}{\ding{51}}%
\newcommand{\xmark}{\ding{55}}%

\newcommand\tool{\textsc{Vieira}}
\newcommand\scallop{\textsc{Scallop}}

\newcommand\numapps{9}

\newcommand\sclsize{\fontsize{8pt}{8pt}\selectfont}

\newcommand\sclsizesmall{\fontsize{7pt}{7pt}\selectfont}

\newcommand\sclsizetiny{\fontsize{6pt}{6pt}\selectfont}

\definecolor{sclgreen}{rgb}{0,0.46,0}
\definecolor{sclblue}{rgb}{0.02,0.42,0.74}
\definecolor{scllightgrey}{rgb}{0.94,0.94,0.94}%
\definecolor{sclgreyblue}{rgb}{0.3,0.4,0.6}%
\definecolor{sclcyan}{rgb}{0.1,0.4,0.6}%
\definecolor{sclpurple}{rgb}{0.71,0,0.85}
\definecolor{sclyellow}{rgb}{0.9,0.6,0.05}
\definecolor{sclorange}{rgb}{1,0.36,0.03}
\definecolor{sclred}{rgb}{0.6,0.2,0.0}%

\definecolor{mygreen}{HTML}{D5E8D4}
\definecolor{myred}{HTML}{F8CECC}
\definecolor{myorange}{HTML}{ffcf99}
\definecolor{myblue}{HTML}{99c8f2}
\definecolor{mypurple}{HTML}{E1D5E7}
\definecolor{myyellow}{HTML}{FFF2CC}

\definecolor{mydeepgreen}{HTML}{82B366}
\definecolor{mydeepred}{HTML}{B85450}
\definecolor{mydeeporange}{HTML}{e38820}
\definecolor{mydeepblue}{HTML}{408bcf}
\definecolor{mydeeppurple}{HTML}{9673A6}
\definecolor{mydeepyellow}{HTML}{D6B656}

\definecolor{mylightblue}{HTML}{cfe2fa}
\definecolor{mylightorange}{HTML}{faeccf}

\lstdefinelanguage{scallopfig}{
    keywords={extern,import,type,case,is,const,rel,query,bound,free,usize,where,as,String,Tensor,DateTime,Duration,i8,i32,i64,usize,u8,u16,u32,u64,f32,f64},%
    keywordstyle=\color{blue},%
    morekeywords=[2]{and,or,not,implies,if,else,then,==,+,-,*,/},%
    keywordstyle=[2]\color{sclpurple},%
    morekeywords=[3]{count,sum,prod,min,max,exists,forall,unique,top,categorical,uniform},%
    keywordstyle=[3]\color{sclorange},%
    morecomment=[s]{/*}{*/},%
    commentstyle=\color{sclgreen},%
    morecomment=[l]{//},%
    morestring=[b]{"},%
    stringstyle=\color{sclyellow},%
}

\lstdefinelanguage{scallop}{
}

\lstdefinelanguage{mypython}{
    keywords={class,def,assert,and,str,return,if,elif,else,for,in,while,yield,zip,int,List,Tuple},keywordstyle=\color{blue},%
    morekeywords=[2]{self},
    keywordstyle=[2]\color{sclred},
    morekeywords=[3]{__init__},
    keywordstyle=[3]\color{sclcyan},
    morecomment=[s]{"""}{"""},commentstyle=\color{sclgreen},%
    morecomment=[l]{\#},%
    morestring=[b]",stringstyle=\color{sclorange}
}

\newcommand{\inlinecode}[1]{{\texttt{\small #1}}}
\newcommand{\inlinescl}[1]{{\small \lstinline[language=scallop]!#1!}}

\begin{document}

\maketitle

\begin{abstract}
Foundation models have vast potential to enable diverse AI applications.
The powerful yet incomplete nature of these models has spurred a wide range of mechanisms to augment them with capabilities such as in-context learning, information retrieval, and code interpreting.
We propose \tool{}, a declarative framework that unifies these mechanisms in a general solution for programming with foundation models.
\tool{} follows a probabilistic relational paradigm and treats foundation models as stateless functions with relational inputs and outputs.
It supports neuro-symbolic applications by enabling the seamless combination of such models with logic programs, as well as complex, multi-modal applications by streamlining the composition of diverse sub-models.
We implement \tool{} by extending the \scallop{} compiler with a foreign interface that supports foundation models as plugins.
We implement plugins for 12 foundation models including GPT, CLIP, and SAM.
We evaluate \tool{} on \numapps{} challenging tasks that span language, vision, and structured and vector databases.
Our evaluation shows that programs in \tool{} are concise, can incorporate modern foundation models, and have comparable or better accuracy than competitive baselines.
\end{abstract}

\section{Introduction}
\label{sec:intro}

Foundation models are deep neural models that are trained on a very large corpus of data and can be adapted to a wide range of downstream tasks \cite{rishi2021foundationmodels}.
Exemplars of foundation models include \textit{language models} (LMs) like GPT \cite{bubeck2023sparks}, \textit{vision models} like Segment Anything \cite{kirillov2023segment}, and \textit{multi-modal models} like CLIP \cite{alec2021transferablevisualmodels}.
While foundation models are a fundamental building block, they are inadequate for programming AI applications end-to-end.
For example, LMs \textit{hallucinate} and produce nonfactual claims or incorrect reasoning chains \cite{mckenna2023sources}.
Furthermore, they lack the ability to reliably incorporate structured data, which is the dominant form of data in modern databases.
Finally, composing different data modalities in custom or complex patterns remains an open problem,
despite the advent of multi-modal foundation models such as ViLT \cite{alec2021transferablevisualmodels} for visual question answering.

Various mechanisms have been proposed to augment foundation models to overcome these limitations.
For example, PAL \cite{gao2023pal}, WebGPT \cite{reiichiro2021webgpt}, and Toolformer \cite{schick2023toolformer} connect LMs with search engines and external tools, expanding their information retrieval and structural reasoning capabilities.
LMQL \cite{beurer2022prompting} generalizes pure text prompting in LMs to incorporate scripting.
In the domain of computer vision (CV), neuro-symbolic visual reasoning frameworks such as \textsc{VisProg} \cite{Gupta2022VisProg} compose diverse vision models with LMs and image processing subroutines.
Despite these advances, programmers lack a general solution that systematically incorporates these methods into a single unified framework.

\begin{figure}
  \footnotesize
  \begin{subfigure}[t]{\linewidth}
    \begin{lstlisting}[language=scallopfig,basicstyle=\sclsizesmall\ttfamily,frame=single,xleftmargin=0.2cm,xrightmargin=0.2cm,framesep=5px,framexleftmargin=0pt,framexrightmargin=0pt]
@gpt("The height of {{x}} is {{y}} in meters")
type height(bound x: String, y: i32)

// Retrieving height of mountains
rel mount_height(m, h) = mountain(m) and height(m, h)
    \end{lstlisting}
    \vspace{-3px}
    \caption{Program {\bf P1}: Extracting knowledge using GPT.}
    \vspace{3px}
    \label{fig:moti-a}
  \end{subfigure}
  \begin{subfigure}[t]{\linewidth}
    \begin{lstlisting}[language=scallopfig,basicstyle=\sclsizesmall\ttfamily,frame=single,xleftmargin=0.2cm,xrightmargin=0.2cm,framesep=5px,framexleftmargin=0pt,framexrightmargin=0pt]
@clip(["cat", "dog"])
type classify(bound img: Tensor, label: String)

// Classify each image as cat or dog
rel cat_or_dog(i, l) = image(i, m) and classify(m, l)
    \end{lstlisting}
    \vspace{-3px}
    \caption{Program {\bf P2}: Classifying images using CLIP.}
        \vspace{3px}
    \label{fig:moti-b}
  \end{subfigure}
  \begin{subfigure}[t]{\linewidth}
    \centering
    \includegraphics[width=\linewidth]{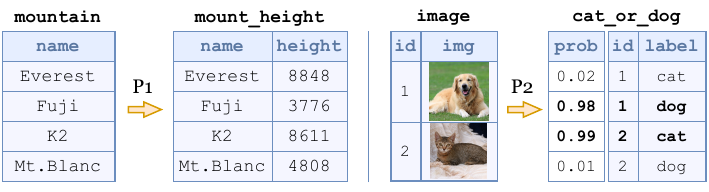}
    \vspace{-10px}
    \caption{Example input-output relations of the programs.}
    \label{fig:moti-c}
  \end{subfigure}
  \caption{Programs in \tool{} using foundation models.}
  \label{fig:motivating-examples}
  \vspace{-10px}
\end{figure}

In this paper, we propose \tool{}, a declarative framework for programming with foundation models.
\tool{} follows a (probabilistic) relational paradigm due to its theoretical and practical versatility.
Structured data is commonly stored in relational databases.
Relations can also represent structures such as \textit{scene graphs} in vision and \textit{abstract syntax trees} in natural and formal languages.
Moreover, extensions for probabilistic and differentiable reasoning enable the integration of relational programming with deep learning in neuro-symbolic frameworks like DeepProbLog \cite{robin2018deepproblog} and \scallop{} \cite{li2023scallop}.

In \tool{}, relations form the abstraction layer for interacting with foundation models.
Our key insight is that foundation models are {\it \textbf{stateless functions} with relational inputs and outputs}.
Fig.~\ref{fig:moti-a} shows a \tool{} program which invokes GPT to extract the height of mountains whose names are specified in a structured table.
Likewise, the program in Fig.~\ref{fig:moti-b} uses the image-text alignment model CLIP to classify images into discrete labels such as \inlinecode{cat} and \inlinecode{dog}.
Fig.~\ref{fig:moti-c} shows relational input-output examples for the two programs.
Notice that the CLIP model also outputs probabilities that allow for probabilistic reasoning.


We implement \tool{} by extending the \scallop{} compiler with a {\it \textbf{foreign interface} that supports foundation models as \textbf{plugins}}.
We implement a customizable and extensible plugin library comprising 12 foundation models including GPT, CLIP, and SAM.
The resulting unified interface enables a wide spectrum of applications with benefits such as reduced hallucination, retrieval augmentation, and multi-modal compositionality.
We evaluate \tool{} on \numapps{} applications that span natural language reasoning, information retrieval, visual question answering, image generation, and image editing.
For these applications, we explore diverse methods for programming with foundation models, such as 
neuro-symbolic reasoning, combining semantic searching with question answering, and modularly composing foundation models.
We not only observe on-par or superior performance of our solutions compared to competitive baselines, but also demonstrate their succinctness and ease-of-use.

We summarize our contributions as follows:
(1) we introduce a new approach based on relational programming to build applications on top of foundation models;
(2) we implement an extensible plugin library of 12 programmable foundation models; and
(3) we evaluate \tool{} on \numapps{} benchmark tasks, and demonstrate comparable or better no-training accuracy than neural-only as well as task-specific baselines.
Our framework, plugin library, and evaluations are open-source and available at \url{https://github.com/scallop-lang/scallop}.

\section{Related Work}
\label{sec:related-work}

\paragraph{Neuro-symbolic methods.}
These methods combine the complementary benefits of neural learning and symbolic reasoning.
They include domain-specific solutions \cite{yi2018nsvqa, mao2019nscl, li2020closed, wang2019satnet, xu2022dont, chen2020nerd, minervini2020ctp} as well as general programming frameworks, such as DeepProbLog \cite{robin2018deepproblog} and \scallop{} \cite{li2023scallop}.
These methods typically concern training or fine-tuning neural models in the presence of logical programs, whereas we target building applications atop foundation models with zero-shot or few-shot examples.
Another recent work, the STAR framework \cite{rajasekharan2023star} also connects a language model (neural) to an answer set programming reasoner (symbolic). 
It is conceptually similar to \tool{} but only focuses on natural language understanding and does not support probabilistic reasoning.



\paragraph{Foundation models.}

These models target different modalities and domains \cite{touvron2023llama, openai2023gpt4, alec2021transferablevisualmodels, kirillov2023segment, alec2021transferablevisualmodels}.
Their reasoning capabilities continue to improve with larger context sizes \cite{ratner2023parallel}, smarter data selection \cite{adadi2021survey}, and the discovery of new prompting methods, such as chain-of-thought \cite{wei2023chainofthought, kojima2022large}, self-consistency \cite{wang2023selfconsistency}, and ReAct \cite{yao2023react}.
\tool{} is orthogonal to these techniques and stands to further enhance the robustness and reliability of foundation models in end-to-end AI applications.

\paragraph{Tools aiding language models.}
There are many efforts that seek to improve the reasoning abilities of language models (LMs) by incorporating external programs and tools \cite{gao2023pal, schick2023toolformer, reiichiro2021webgpt, davis2023testing}.
For instance, AutoGPT \cite{autogpt} and TaskMatrix.AI \cite{liang2023taskmatrixai} allows black-box LMs to control symbolic reasoning by invoking commands or calling APIs.
On the other hand, many works attempt to extract structured information from LMs for downstream tasks \cite{Gupta2022VisProg, beurer2022prompting}.
\tool{} unifies these two strategies for augmenting model capabilities, and extends them into a glue language for composing multi-modal foundation models.












\section{Language}
\label{sec:language}

\tool{} employs a declarative logic programming language based on Datalog \cite{alice}.
In this section, we present the core language and its \textit{foreign interface} for incorporating diverse foundation models.


\subsection{Core Language}
\label{sec:language-core}

\paragraph{Relations and data types.}

The fundamental data type in \tool{} is set-valued relations comprising tuples of statically-typed primitive values.
Besides the standard primitive types such as integers (e.g. \inlinecode{i32}) and string (\inlinecode{String}), \tool{} introduces two additional types for seamless integration of foundation models: \inlinecode{Tensor} and \textit{Algebraic Data Types} (ADTs).
For example, we can declare a relation named \inlinecode{image} to store tuples of image IDs and image  \inlinecode{Tensor}s:
\begin{lstlisting}[language=scallop]
type image(img_id: i32, img: Tensor)
\end{lstlisting}
The contents of this relation can be specified via a set of tuples using the built-in \textit{foreign function} \inlinecode{\$load\_image}:
\begin{lstlisting}[language=scallop]
rel image = {(0, $load_image("cat.png")), ...}
\end{lstlisting}
ADTs in \tool{} enable the specification of domain specific languages (DSLs) to bridge structured and unstructured data.
For example, the following DSL for visual question answering (VQA) describes queries to retrieve scene objects, count objects, and check the existence of objects:
\begin{lstlisting}[language=scallop]
type Query = Scene() | Filter(Query, String) 
           | Count(Query) | Exists(Query) | ...
// How many balls are there?
const MY_QUERY = Count(Filter(Scene(), "ball"))
\end{lstlisting}

\paragraph{Logical reasoning.}

Being based on Datalog, \tool{} supports defining Horn rules, thereby allowing logical reasoning constructs such as conjunction, disjunction, recursion, stratified negation, and aggregation.
Recursion is particularly useful for inductively defining the semantics of a DSL.
For example, a (partial) semantics for the above DSL is defined as follows, where \inlinecode{eval\_o} and \inlinecode{eval\_n} are recursively defined to evaluate objects and numbers, respectively:
\begin{lstlisting}[language=scallop]
// Scene returns all objects
rel eval_o(e, o) = case e is Scene() and obj(o)
// Filter applies filter using attributes
rel eval_o(e, o) = case e is Filter(f, a)
    and eval_o(f, o) and attr(o, a)
// Count returns the number of evaluated objects
rel eval_n(e, n) = n := count(o: eval_o(e1, o) 
    where e1: case e is Count(e1))
... // other cases of `e'
\end{lstlisting}
Note that the \inlinecode{case}-\inlinecode{is} operator matches patterns of the ADT and the \inlinecode{count} aggregator counts the number of entities.
When combined with foundation models, principled reasoning semantics in this style can compensate for individual foundation models' lack of reasoning capability.

\paragraph{Probabilistic soft logic.}
Tuples can be tagged with probabilities.
The example below shows hard-coded probabilities, suggesting that the entity is more likely a dog than a cat:
\begin{lstlisting}[language=scallop]
rel animal = {0.1::(1,"cat"), 0.9::(1,"dog")}
\end{lstlisting}
Soft-logic operations produce probabilities as well.
For instance, the \textit{soft-eq} operator (\inlinecode{\~=}) on \inlinecode{Tensor}s derives cosine-similarity between tensors, enabling features like \textit{soft-join} and applications like \textit{semantic search}.
In the following example, we compute similarity scores between distinct documents by performing \textit{soft-join} on their embeddings:
\begin{lstlisting}[language=scallop]
type doc(id: i32, embed: Tensor) // embed docs
rel sim(i, j) = doc(i, v) and doc(j, v) and i!=j
// equiv: sim(i, j) = doc(i, v1) and doc(j, v2) and i!=j and v1~=v2
\end{lstlisting}
Notice that in the above rule, a join on a tensor value \inlinecode{v} is de-sugared into a soft-eq on two individual variables (denoted \inlinecode{v1} and \inlinecode{v2}).
Internally, with the provenance framework provided by \scallop{} \cite{li2023scallop}, we use the top-$k$-proofs semiring \cite{huang2021scallop} for scalable probabilistic reasoning, thus enabling features such as ranking and uncertainty estimation.

\subsection{Foreign Interface}
\label{sec:language-fi}

In order to incorporate foundation models, we design a foreign interface with two main programming constructs, called \textit{foreign predicate} and \textit{foreign attribute}.
They can be defined externally in languages like Python and imported into \tool{} for application.




\paragraph{Foreign Predicate (FP).}

Foreign predicates can be used in rules just like other relations.
However, instead of grounding relational facts from a table, FPs ground facts by invoking external functions.
The syntax for defining FPs is as follows:
\begin{lstlisting}[language=scallop]
extern type PRED([bound|free]? ARG: TYPE, ...)
\end{lstlisting}
In addition to the type, each argument is specified either as a \textit{bounded} argument (using the keyword \inlinecode{bound}) or a \textit{free} argument (using \inlinecode{free} or omitted for brevity).
Semantically, FPs are functions that take in a tuple of bounded arguments and return a list of tuples of free arguments.
The runtime of \tool{} performs memoization on FP results to avoid redundant computation.
Optionally, FPs can tag a probability to each returned tuple for further probabilistic reasoning.


\paragraph{Foreign Attribute (FA).}

\begin{figure}
\begin{lstlisting}[language=mypython,frame=single,framesep=6px,xrightmargin=5px]
@foreign_attribute
def clip(pred: Predicate, labels: List[str]):
  # Sanity checks for predicate and labels...
  assert pred.args[0].ty == Tensor and ...
  
  @foreign_predicate(name=pred.name)
  def run_clip(img: Tensor) -> Facts[str]:
    # Invoke CLIP to classify image into labels
    probs = clip_model(img, labels)
    # Each result is tagged by a probability
    for (prob, label) in zip(probs, labels):
      yield (prob, (label,)) # prob::(label,) 
      
  return run_clip
\end{lstlisting}
\caption{
Snippet of Python implementation of the foreign attribute \inlinecode{clip} which uses the CLIP model for image classification.
Notice that the FA \inlinecode{clip} returns the FP \inlinecode{run\_clip}.
}
\label{fig:clip-fa}
\vspace{-5px}
\end{figure}

In \tool, attributes can be used to \textit{decorate} declarations of predicates.
They are higher-order functions that take in the provided arguments and the decorated predicate to return a new predicate.
The syntax for using an attribute to decorate a predicate is:
\begin{lstlisting}[language=scallop,numbers=none]
@ATTR(POS_ARG, ..., KEY=KW_ARG, ...)
type PRED([bound|free]? ARG: TYPE, ...)
\end{lstlisting}
The attribute is applied prior to the compilation of \tool{} programs.
For interfacing with foundation models, the positional and keyword arguments are particularly helpful in configuring the underlying model, hiding low-level details.
Fig.~\ref{fig:clip-fa} illustrates one succinct implementation of the FA that enables the use of the CLIP model shown in Fig. \ref{fig:moti-b}.




\section{Foundation Models}
\label{sec:foundation-models}


\tool{} provides an extensible \textit{plugin framework} that adapts to the evolving landscape of foundation models.
In this work, we have implemented 7 plugins, covering 12 foundation models, all through the foreign interface.
Our design principle for the interface is three-fold: simplicity, configurability, and compositionality.
In this section, we present several representative predicates and attributes which substantially support the applicability of \tool{} to diverse machine learning tasks.

\paragraph{Text completion.}

In \tool, language models like GPT \cite{openai2023gpt4} and LLaMA \cite{touvron2023llama} can be used as basic foreign predicates for text completion:
\begin{lstlisting}[language=scallop,numbers=none]
extern type gpt(bound p: String, a: String)
rel ans(a) = gpt("population of NY is", a)
\end{lstlisting}
In this case, \inlinecode{gpt} is an arity-2 FP that takes in a \inlinecode{String} as the prompt and produces a \inlinecode{String} as the response.
It uses the model \inlinecode{gpt-3.5-turbo} by default.
To make the interface more relational and structural, we provide an FA:
\begin{lstlisting}[language=scallop,numbers=none]
@gpt("the population of {{loc}} is {{num}}",
  examples=[("NY", 8468000), ...])
type population(bound loc: String, num: u32)
\end{lstlisting}
Here, we declare a relation named \inlinecode{population} which produces a population number (\inlinecode{num}) given a location (\inlinecode{loc}) as input.
Notice that structured few-shot examples are provided through the argument \inlinecode{examples}.

\paragraph{Semantic parsing.}

One can directly configure language models to perform \textit{semantic parsing}.
For instance, the semantic parser for the simple \inlinecode{Query} DSL (partially defined in the Language section) can be declared as follows:
\begin{lstlisting}[language=scallop,numbers=none]
@gpt_semantic_parse(
  "Please semantically parse questions...",
  examples=[("How many red things are there?",
    "Count(Filter(Scene(), 'red'))"), ...])
type parse_query(bound x: String, y: Query)
\end{lstlisting}
Internally, the language model is expected to generate a fully structured \inlinecode{Query} in its string form. 
Then, \tool{} attempts to parse the string to construct actual ADT values.
In practice, the success of semantic parsing depends heavily on the design of the DSL, involving factors like intuitiveness (e.g., names and arguments of ADT variants) and complexity (e.g., number of possible ADT variants).



\paragraph{Relational data extraction.}

Structural relational knowledge available in free-form textual data can be extracted by language models.
We introduce a foreign attribute \inlinecode{@gpt\_extract\_relation} for this purpose.
For instance, the following declared predicate takes in a context and produces (subject, object, relation) triplets:
\begin{lstlisting}[language=scallop,numbers=none]
@gpt_extract_relation(
  "Extract the implied kinship relations",
  examples=[("Alice and her son Bob went to...",
    [("alice", "bob", "son"), ...])])
type extract_kinship(bound ctx: String,
  sub: String, obj: String, rela: String)
\end{lstlisting}
This attribute differs from the text completion attribute in that it can extract an arbitrary number of facts.
The underlying implementation prompts LMs to respond with JSON-formatted strings, allowing structured facts to be parsed.

\paragraph{Language models for textual embedding.}

Textual embeddings are useful in performing tasks such as information retrieval.
The following example declares an FP encapsulating a cross-encoder \cite{nogueira2019passage}:
\begin{lstlisting}[language=scallop,numbers=none]
@cross_encoder("nli-deberta-v3-xsmall")
type enc(bound input: String, embed: Tensor)
rel sim() = enc("cat", e) and enc("neko", e)
\end{lstlisting}
In the last line, we compute the cosine-similarity of the encoded embeddings using a soft-join on the variable \inlinecode{e}.
As a result, we obtain a probabilistic fact like \inlinecode{0.9::sim()} whose probability encodes the cosine-similarity between the textual embeddings of \inlinecode{"cat"} and \inlinecode{"neko"}.

\paragraph{Image classification models.}

Image-text alignment models, such as CLIP \cite{alec2021transferablevisualmodels}, can naturally be used as zero-shot image classification models.
Fig.~\ref{fig:moti-b} shows an example usage of the \inlinecode{@clip} attribute.
We also note that dynamically-generated classification labels can be provided to CLIP via a bounded argument in the predicate.

\paragraph{Image segmentation models.}

OWL-ViT \cite{minderer2022simple}, Segment Anything Model (SAM) \cite{kirillov2023segment}, and DSFD \cite{jian2018dsfd} are included in \tool~as image segmentation (IS) and object localization (LOC) models.
IS and LOC models can provide many outputs, such as bounding boxes, classified labels, masks, and cropped images.
For instance, the OWL-ViT model can be used and configured as follows:
\begin{lstlisting}[language=scallop,numbers=none]
@owl_vit(["human face", "rocket"])
type find_obj(bound img: Tensor,
  id: u32, label: String, cropped_image: Tensor)
\end{lstlisting}
Here, the \inlinecode{find\_obj} predicate takes in an image, and finds image segments containing ``human face'' or ``rocket''.
According to the names of the arguments, the model extracts 3 values per segment: ID, label, and cropped image.
Note that each produced fact will be associated with a probability, representing the confidence from the model.

\paragraph{Image generation models.}

Visual generative models such as Stable Diffusion \cite{Rombach_2022_diffusion} and DALL-E \cite{ramesh2021zero} can be regarded as relations as well.
The following example shows the declaration of the \inlinecode{gen\_image} predicate, which encapsulates a diffusion model:
\begin{lstlisting}[language=scallop,numbers=none]
@stable_diffusion("stable-diffusion-v1-4")
type gen_image(bound txt: String, img: Tensor)
\end{lstlisting}
As can be seen from the signature, it takes in a \inlinecode{String} text as input and produces a \inlinecode{Tensor} image as output.
Optional arguments such as the desired image resolution and the number of inference steps can be supplied to dictate the granularity of the generated image.

\section{Tasks and Solutions}
\label{sec:tasks}

\begin{table}[bt!]
\footnotesize
\centering
\setlength{\tabcolsep}{0.10cm}
\begin{tabular}{
>{\centering\arraybackslash}m{1.5cm}|
>{\centering\arraybackslash}m{1.5cm}
>{\centering\arraybackslash}m{1.3cm}|
c|
>{\centering\arraybackslash}m{1.8cm}}
\toprule
\textbf{Task} 
    & \textbf{Dataset}                      
    & \textbf{\#Test Samples}        
    & \textbf{Metric} 
    & \textbf{Foundation Models Used}      
\\ \midrule \midrule
DR               
    & DR                    
    & 369      
    & EM 
    & GPT-4
\\ \hline
TSO   
    & TSO              
    & 150   
    & EM
    & GPT-4
\\ \hline
KR        
    & CLUTRR                       
    & 1146  
    & EM
    & GPT-4 
\\ \hline
MR              
    & GSM8K                       
    & 1319  
    & EM
    & GPT-4
\\ \hline
\multirow{2}{1.5cm}{\centering QA}  
    & \multirow{2}{*}{Hotpot QA}   
    & \multirow{2}{*}{1000} 
    & \multirow{2}{1cm}{\centering EM} 
    & GPT-4
    \\ \cline{5-5}
    &                              
    &                       
    & 
    &  ada-002                       
\\ \hline
\multirow{2}{1.5cm}{\centering PS}  
    & \multirow{2}{1.5cm}{\centering Amazon ESCI} 
    & \multirow{2}{*}{1000} 
    & \multirow{2}{*}{nDCG}  
    & GPT-4
    \\ \cline{5-5}
    &                              
    &                       
    &                        
    & ada-002
\\ \hline
\multirow{4}{1.5cm}{\centering VQA} 
    & \multirow{2}{*}{CLEVR}   
    & \multirow{2}{*}{480}
    & \multirow{4}{1.2cm}{\centering Recall@1 Recall@3}
    & GPT-4
    \\ \cline{5-5} 
    &
    &
    &
    & OWL-ViT 
    \\ \cline{2-3}\cline{5-5}
    & \multirow{2}{*}{GQA}         
    & \multirow{2}{*}{500}     
    &          
    & VilT                   
    \\ \cline{5-5}
    &                              
    &                       
    &                        
    & CLIP
\\ \hline
\multirow{5}{1.5cm}{\centering VOT}
    & \multirow{3}{*}{VQAR}        
    & \multirow{3}{*}{100}  
    & \multirow{5}{*}{MI}    
    & OWL-ViT 
    \\ \cline{5-5}
    &                              
    &                       
    &           
    & VilT    
    \\ \cline{5-5}
    &
    &
    &
    & GPT-4
    \\ \cline{2-3}\cline{5-5}
    & \multirow{2}{*}{OFCP}
    & \multirow{2}{*}{50}  
    & 
    & DSFD
    \\ \cline{5-5}
    &
    &
    &
    & CLIP
\\ \hline
\multirow{4}{1.5cm}{\centering IGE}  
    & \multirow{2}{1.5cm}{\centering OFCP} 
    & \multirow{2}{*}{50} 
    & \multirow{4}{*}{MI}  
    & DFSD 
    \\ \cline{5-5}
    &                              
    &                       
    &           
    & CLIP
    \\ \cline{2-3}\cline{5-5}
    & \multirow{2}{1.5cm}{\centering IGP20} 
    & \multirow{2}{*}{20}   
    &           
    & GPT-4
    \\ \cline{5-5}
    &
    &
    &
    & Diffusion
\\ \hline
\bottomrule
\end{tabular}
\caption{
Characteristics of benchmark tasks including the dataset used, its size, and evaluation metrics. Metrics include exact match (EM), normalized discounted cumulative gain (nDCG), and manual inspection (MI).
We also denote the foundation models used in our solution for each task.
}
\label{tab:task-spec}
\end{table}

\begin{figure*}[tb!]
    \centering
    \footnotesize
    \includegraphics[width=\linewidth]{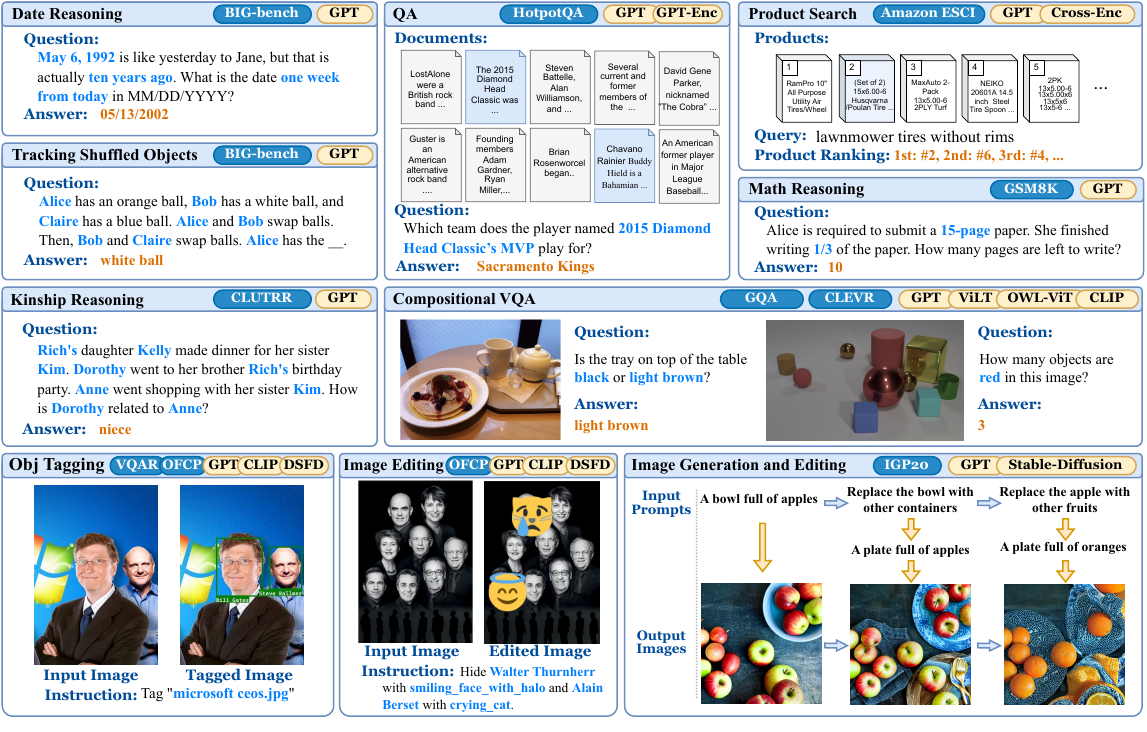}
    \vspace{-20px}
    \caption{
        Benchmark tasks.
        The top of each box lists the dataset(s) and the foundation models used in our solutions.
    }
    \label{fig:benchmark-tasks}
\end{figure*}

We apply \tool{} to solve \numapps{} benchmark tasks depicted in Fig.~\ref{fig:benchmark-tasks}.
Table \ref{tab:task-spec} summarizes the datasets, evaluation metrics, and the foundation models used in our solutions.
We elaborate upon the evaluation settings and our solutions below. 

\paragraph{Date reasoning (DR).}

In this task adapted from BIG-bench \cite{srivastava2023imitation}, the model is given a context and asked to compute a date.
The questions test the model's temporal and numerical reasoning skills, as well as its grasp of common knowledge.
Unlike BIG-bench where multiple-choice answers are given, we require the model to directly produce its answer in MM/DD/YYYY form.

Our solution leverages GPT-4 (5-shot\footnote{In this work, $k$ in ``$k$-shot'' means the number of examples provided to the LM component within the full solution. Each example is a ground-truth input-output pair for the LM.}) for extracting 3 relations: mentioned dates, duration between date labels, and the target date label.
From here, our relational program iterates through durations to compute dates for all date labels.
Lastly, the date of the target label is returned as the output.

\paragraph{Tracking shuffled objects (TSO).}

In this task from BIG-bench, a textual description of pairwise object swaps among people is given, and the model needs to track and derive which object is in a specified person's possession at the end.
There are three difficulty levels depending on the number of objects to track, denoted by $n \in \{3, 5, 7\}$.

Our solution for tracking shuffled objects relies on GPT-4 (1-shot) to extract 3 relations: initial possessions, swaps, and the target person whose final possessed object is expected as the answer.
Our reasoning program iterates through all the swaps starting from the initial state and retrieves the last possessed object associated with the target.

\paragraph{Kinship reasoning (KR).}

CLUTRR \cite{sinha2019clutrr}~is a kinship reasoning dataset of stories which indicate the kinship between characters, and requires the model to infer the relationship between two specified characters.
The questions have different difficulty levels based on the length of the reasoning chain, denoted by $k \in \{2 \dots 10\}$.


Our solution for kinship reasoning invokes GPT-4 (2-shot) to extract the kinship graph from the context.
We also provide an external common-sense knowledge base for rules like ``mother's mother is grandmother''.
Our program then uses the rules to derive other kinship relations.
Lastly, we retrieve the kinship between the specified pair of people.

\paragraph{Math reasoning (MR).}

This task is drawn from the GSM8K dataset of arithmetic word problems \cite{cobbe2021training}.
The questions involve grade school math word problems created by human problem writers, and the model is asked to produce a number as the result.
Since the output can be fractional, we allow a small delta when comparing the derived result with the ground truth.

Our solution to this task prompts GPT-4 (2-shot) to produce step-by-step expressions, which can contain constants, variables, and simple arithmetic operations.
We evaluate all the expressions through a DSL, and the result associated with the goal variable is returned.
By focusing the LM's responsibility solely on semantic parsing, our relational program can then achieve faithful numerical computation via DSL evaluation.

\paragraph{Question answering with information retrieval (QA).}

We choose HotpotQA \cite{yang2018hotpotqa}, a Wikipedia-based question answering (QA) dataset under the ``distractor'' setting.
Here, the model takes in 2 parts of inputs:
1) a question, and
2) 10 Wikipedia paragraphs as the context for answering the question.
Among the 10 Wikipedia pages, at most 2 are relevant to the answer, while the others are distractors.

Our solution is an adaptation of FE2H \cite{li2022fe2h}, which is a 2-stage procedure.
First, we turn the 10 documents into a vector database by embedding each document.
We then use the embedding of the question to retrieve the 2 most related documents, which are then fed to a language model to do QA.
In this case, the QA model does not have to process all 10 documents, leading to less distraction.

\paragraph{Product search (PS).}

We use Amazon's ESCI Product Search dataset \cite{reddy2022shopping}.
The model is provided with a natural language (NL) query and a list of products (23 products on average).
The goal is to rank the products that best match the query.
In the dataset, for each pair of query and product, a label among $E$ (exact match), $S$ (substitute), $C$ (complementary), and $I$ (irrelevant) is provided.
The metric we use to evaluate the performance is nDCG.
The gains are set to be 1.0 for $E$, 0.1 for $S$, 0.01 for $C$, and 0.0 for $I$.

One challenge of this dataset is that many queries contain negative statements.
For example, in the query ``\#1 treadmill without remote'', the ``remote'' is undesirable.
Therefore, instead of computing the embedding of the full query, we decompose the query into positive and negative parts.
We then perform semantic search by maximizing the similarity of the positive part while minimizing that of the negative part.

\paragraph{Compositional visual question answering (VQA).}

We choose two compositional VQA datasets, GQA \cite{drew2019gqa} and CLEVR \cite{johnson2016clevr}.
In this task, the model is given an image and a question, and needs to answer the question.
For GQA, the majority of questions expect yes/no answers, while CLEVR's questions demand features like counting and spatial reasoning.
We uniformly sample 500 and 480 examples from GQA and CLEVR datasets respectively.
Following VQA conventions \cite{kim2021vilt}, we use Recall@$k$ where $k \in \{1, 3\}$ as the evaluation metrics.

Our solution for GQA is an adaptation of \textsc{VisProg} \cite{Gupta2022VisProg}.
We create a DSL for invoking vision modules such as ViLT and OWL-ViT, and use GPT-4 for converting questions into programs in this DSL.
Our solution for CLEVR is similar, directly replicating the DSL provided by the original work.
OWL-ViT and CLIP are used to detect objects and infer attributes, while the spatial relations are directly computed using the bounding box data.

\paragraph{Visual object tagging (VOT).}

We evaluate on two datasets, VQAR \cite{huang2021scallop} and OFCP.
For VQAR, the model is given an image and a programmatic query, and is asked to produce bounding boxes of the queried objects in the image.
Our solution composes a relational knowledge base, defining entity names and relationships, with object retrieval (OWL-ViT) and visual QA (ViLT) models.

Online Faces of Celebrities and Politicians (OFCP) is a self-curated dataset of images from Wikimedia Commons among other sources.
For this dataset, the model is given an image with a descriptive NL filename, and needs to detect faces relevant to the description and tag them with their names. 
Our solution obtains a set of possible names from GPT-4 and candidate faces from DSFD. 
These are provided to CLIP for object classification, after which probabilistic reasoning filters the most relevant face-name pairs.

\paragraph{Language-guided image generation and editing (IGE).}

We adopt the task of image editing from \cite{Gupta2022VisProg}.
In this task, the instruction for image editing is provided through NL, and can invoke operations such as blurring background, popping color, and overlaying emojis.
Due to the absence of an existing dataset, we repurpose the OFCP dataset by introducing 50 NL image editing prompts. 
Our solution for this task is centered around a DSL for image editing.
We incorporate GPT-4 for semantic parsing, DSFD for face detection, and CLIP for entity classification.
Modules for image editing operations are implemented as individual foreign functions.

For free-form generation and editing of images, we curate IGP20, a set of 20 prompts for image generation and editing.
Instead of using the full prompt, we employ an LM to decompose complex NL instructions into simpler steps.
We define a DSL with high-level operators such as generate, reweight, refine, replace, and negate.
We use a combination of GPT-4, Prompt-to-Prompt \cite{hertz2022prompttoprompt}, and diffusion model \cite{Rombach_2022_diffusion} to implement the semantics of our DSL.
We highlight our capability of grounding positive terms from negative phrases, which enables handling prompts like ``replace apple with other fruits'' (Fig. \ref{fig:benchmark-tasks}).



\section{Experiments and Analysis}
\label{sec:evaluation}


\begin{table}
  \centering
  \footnotesize
  \setlength{\tabcolsep}{4pt}
  \begin{tabular}{
  >{\centering\arraybackslash}m{1.2cm}
  >{\centering\arraybackslash}m{0.5cm}
  >{\centering\arraybackslash}m{1cm}
  c
  >{\centering\arraybackslash}m{1.8cm}
  >{\centering\arraybackslash}m{0.5cm}
  >{\centering\arraybackslash}m{1cm}}
    \toprule
    \textbf{Dataset} 
    & \textbf{LoC} 
    & \textbf{Prompt LoC} 
    &&
    \textbf{Dataset} 
    & \textbf{LoC} 
    & \textbf{Prompt LoC}
    \\ \cmidrule{1-3} \cmidrule{5-7}
    DR & 69 & 48 & & CLEVR & 178 & 45\\
    TSO & 34 & 16 & & GQA & 82 & 36 \\ 
    CLUTRR & 61 & 45 && VQAR & 53 & 11 \\ 
    GSM8K & 47 & 28 && OFCP (VOT) & 33 & 2 \\
    HotpotQA & 47 & 24 && OFCP (IGE) & 117 & 44 \\ 
    ESCI & 32 & 7 && IGP20 & 50 & 12 \\ 
    \bottomrule
  \end{tabular}
  \caption{
    The lines-of-code (LoC) numbers of our solutions for each dataset.
    The LoC includes empty lines, comments, natural language prompts, and DSL definitions.
    We note specifically the LoC of prompts in the table.
  }
  \label{tab:succinctness}
\end{table}

\begin{table}
  \centering
  \footnotesize
  \setlength{\tabcolsep}{2pt}
  \begin{tabular}{
  l
  >{\centering\arraybackslash}m{1.5cm}
  >{\centering\arraybackslash}m{1.5cm}
  >{\centering\arraybackslash}m{1.5cm}
  >{\centering\arraybackslash}m{1.5cm}}
    \toprule
    \textbf{Method} & 
    \textbf{DR} &
    \textbf{TSO} & 
    \textbf{CLUTRR} & 
    \textbf{GSM8K}
    \\ \midrule
    GPT-4  &
    71.00 (0-shot) &
    30.00 (0-shot) & 
    43.10 (3-shot) &
    87.10 (0-shot)
    \\
    GPT-4 (CoT) & 
    87.26 (0-shot)& 
    84.00 (0-shot) &
    24.17 (3-shot)& 
    \textbf{92.00} (5-shot) \\
    \midrule
    Ours & 
    \textbf{92.41} &
    \textbf{100.00} &
    \textbf{72.50} & 
    90.60
    \\
    \bottomrule
  \end{tabular}
  \caption{
    The performance on the natural language reasoning datasets. 
    Numbers are in percentage (\%).
  }
  \label{tab:nlr-performance}
\end{table}

We aim to answer the following research questions:

\begin{enumerate}
    \setlength{\itemindent}{1.7em}
    \item[\textbf{RQ1.}] Is \tool{} programmable enough to be applicable to a diverse range of applications with minimal effort?
    \item[\textbf{RQ2.}] How do solutions using \tool{} compare to other baseline methods in the no-training setting?
\end{enumerate}

\begin{table}[t!]
  \centering
  \footnotesize
  \setlength{\tabcolsep}{3pt}
  \begin{tabular}{c>{\centering\arraybackslash}m{1.5cm}
  c|c>{\centering\arraybackslash}m{1.5cm}c}
    \toprule
    \multicolumn{3}{c|}{\textbf{HotpotQA}} & \multicolumn{3}{c}{\textbf{Amazon ESCI}} 
    \\ \midrule 
    Method & Fine-tuned & EM & Method & Fine-tuned & nDCG 
    \\ \midrule 
    C2FM & \cmark & 72.07\% & BERT & \cmark & 0.830 \\ 
     FE2H  & \cmark & 71.89\% & CE-MPNet & \cmark & 0.857 \\ \midrule
    --- & --- & --- & MIPS & \xmark & 0.797  \\
    \midrule
    Ours &
    \xmark &
    67.3\% &
    Ours & 
    \xmark & 
    0.798 
    \\
    \bottomrule
  \end{tabular}
  \caption{
  The performance on the HotpotQA and Amazon ESCI.
  We also include performance numbers from methods which are fine-tuned on the corresponding dataset.
  }
  \label{tab:amazon_prod_performance}
\end{table}

\subsection{RQ1: Programmability}

While a user study for \tool{}'s programmability is out of scope in this paper, we qualitatively evaluate its programmability on three aspects.
First, we summarize the lines-of-code (LoC) for each of our solutions in Table \ref{tab:succinctness}.
The programs are concise, as most are under 100 lines.
Notably, natural language prompts (including few-shot examples) take up a significant portion of each solution.
Secondly, 8 out of 10 solutions are coded by undergraduate students with no background in logic and relational programming, providing further evidence of \tool{}'s user-friendliness.
Last but not least, our solutions are interpretable and thus offer debuggability.
Specifically, all the intermediate relations are available for inspection, allowing systematic error analysis.

\subsection{RQ2: Baselines and Comparisons}

\begin{figure}[ht!]
  \centering
  \includegraphics[width=\linewidth]{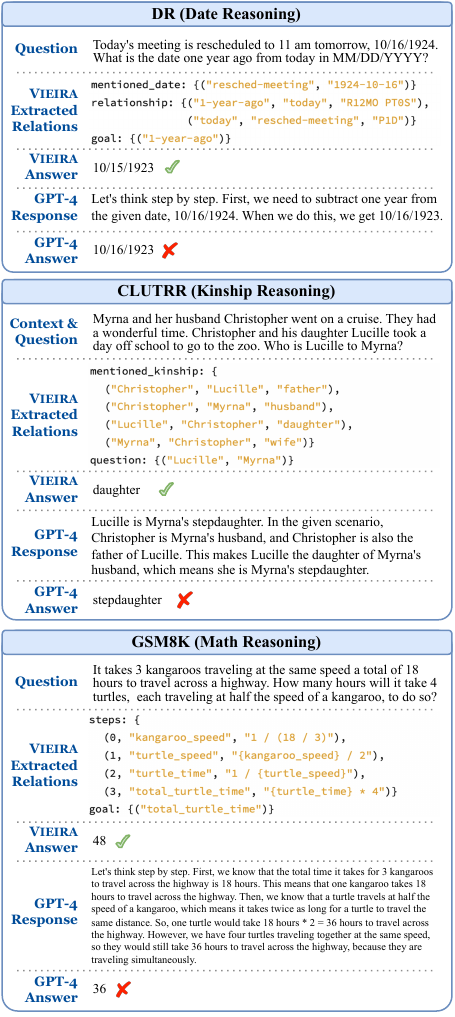}
  \vspace{-10px}
  \caption{
    Illustrative comparisons between our solution and GPT-4 (zero-shot CoT) on selected questions from DR, CLUTRR, and GSM8K datasets.
    We also include the extracted relations used for subsequent reasoning.
  }
  \label{fig:nlr-qualitative-examples}
  \vspace{-10px}
\end{figure}

We compare the performance of our solutions to existing baselines under the no-training setting.
In particular, our solutions achieve better performance than comparable baselines on 6 out of 8 studied datasets with baselines.
Below, we classify the tasks into 4 categories and discuss the respective performance and comparisons.

\paragraph{Natural language reasoning.}

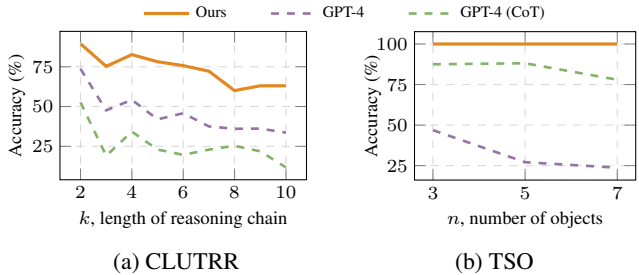
\begin{figure}[t!]
\scriptsize
\centering
\begin{subfigure}[b]{0.55\linewidth}
\scriptsize
\begin{tikzpicture}
    \begin{axis}[
        width=1.05\linewidth, 
        height=1.40in,
        grid=major, 
        grid style={dashed,gray!30}, 
        xlabel={$k$, length of reasoning chain},
        ylabel=Accuracy (\%),
        ytick={25,50,75,100},
        legend style={at={(0.3,1)},style={/tikz/every even column/.append style={column sep=0.5cm}},draw=none,anchor=south west, font=\tiny}, 
        x label style={at={(axis description cs:0.5,0.1)},anchor=north},
        y label style={at={(axis description cs:0.2,.5)}},
        no markers,
        every axis plot/.append style={line width=1pt},
        legend columns=3,
    ]

    \addlegendentry{Ours};
    \addplot[mydeeporange, style=very thick] table[x=k, y=Ours, col sep=comma, ] {data/clutrr-sys-generalizability.csv};
    \addlegendentry{GPT-4};
    \addplot[mydeeppurple, dashed] table[x=k, y=GPT-4, col sep=comma, ] {data/clutrr-sys-generalizability.csv};
    \addlegendentry{GPT-4 (CoT)};
    \addplot[mydeepgreen, dashed] table[x=k, y=GPT-4 (COT), col sep=comma, ] {data/clutrr-sys-generalizability.csv};
    
    \end{axis}
\end{tikzpicture}
\caption{CLUTRR}
\end{subfigure}
\hfill
\begin{subfigure}[b]{0.43\linewidth}
\begin{tikzpicture}
    \begin{axis}[
        width=1.25\linewidth, 
        height=1.40in,
        grid=major, 
        grid style={dashed,gray!30}, 
        xlabel={$n$, number of objects},
        ylabel=Accuracy (\%),
        ytick={25,50,75,100},
        xtick={3,5,7},
        x label style={at={(axis description cs:0.5,0.1)},anchor=north},
        y label style={at={(axis description cs:0.23,.5)}},
        no markers,
        every axis plot/.append style={line width=1pt},
        legend columns=3,
    ]

    \addplot[mydeeporange, style=very thick] table[x=k, y=Ours, col sep=comma, ] {data/TSO-sys-generalizability.csv};
    \addplot[mydeeppurple, dashed] table[x=k, y=GPT-4, col sep=comma, ] {data/TSO-sys-generalizability.csv};
    \addplot[mydeepgreen, dashed] table[x=k, y=GPT-4 (COT), col sep=comma, ] {data/TSO-sys-generalizability.csv};
    
    \end{axis}
\end{tikzpicture}
\caption{TSO}
\end{subfigure}
\vspace{-7px}
\footnotesize
\caption{
Systematic generalizability comparisons on the CLUTRR and TSO datasets.
}
\label{fig:generalizability}
\end{figure}

For the tasks of DR, TSO, CLUTRR, and GSM8K, we pick a generic baseline of GPT-4 under zero-shot, few-shot, and chain-of-thought (CoT) settings.
All our solutions also rely on GPT-4 (few-shot), but we note that our shots only include extracted facts, and not the final answer or any reasoning chains.
The data in Table \ref{tab:nlr-performance} indicates that our method can significantly enhance reasoning performance and reduce hallucination, exemplified by achieving a flawless 100\% accuracy on the TSO dataset.
Note that on GSM8K, our method scores slightly lower than the baseline; we conjecture that our solution demands more from GPT-4 itself to extract structured computation steps.
On CLUTRR, our solution even outperforms fCoT \cite{lyu2023faithful}, a special prompting technique with external tool use, by 0.6\%.
In Fig. \ref{fig:generalizability} we illustrate the systematic generalizability of our methods.
The performance of our solutions remains relatively consistent even when the problems become harder.
We provide illustrative examples in Fig. \ref{fig:nlr-qualitative-examples} showing comparisons between our method and GPT-4 (zero-shot CoT).

\paragraph{Retrieval augmentation and semantic search.}

For the HotpotQA dataset, our solution is an adaptation of FE2H \cite{li2022fe2h}, a retrieval-augmented question answering approach.
As seen in Table \ref{tab:amazon_prod_performance}, with no fine-tuning, our method scores only a few percentages lower than fine-tuned methods C2FM \cite{yin2022c2fm} and FE2H.
For the Amazon ESCI dataset, our solution performs semantic search for product ranking.
While performing slightly lower than the fine-tuned methods \cite{reddy2022shopping, song2020mpnet}, our solution outperforms maximum inner product search (MIPS) based on GPT text encoder (\inlinecode{text-embedding-ada-002}).



\begin{table}[t!]
\centering
\footnotesize
  \setlength{\tabcolsep}{5pt}
  \begin{tabular}{lcccccc}
    \toprule
    \multirow{2}{*}{\textbf{Method}} & 
    \multicolumn{2}{c}{\textbf{GQA}} & 
    \multicolumn{2}{c}{\textbf{CLEVR}}
    \\ \cmidrule{2-3} \cmidrule{4-5}
    &
    Recall@1 &
    Recall@3 &
    Recall@1 &
    Recall@3
    \\ \midrule
    ViLT-VQA & 0.049 & 0.462 & 0.241 & 0.523 \\ 
    PNP-VQA & 0.419 & --- & --- & --- \\
    \midrule
    Ours & 
    \textbf{0.579} &
    \textbf{0.665} & 
    \textbf{0.463} &
    \textbf{0.638}
    \\
    \bottomrule
  \end{tabular}
\captionof{table}{
Quantitative results on the VQA datasets.
}
\label{tab:vqa_performance}
\end{table}

\paragraph{Compositional multi-modal reasoning.}

For VQA, we pick ViLT-VQA \cite{kim2021vilt} (a pre-trained foundation model) and PNP-VQA \cite{tiong-2022-pnpvqa} (a zero-shot VQA method) as baselines.
As shown in Table \ref{tab:vqa_performance}, our method significantly outperforms the baseline model on both datasets.
Compared to the neural-only baseline, our approach that combines DSL and logical reasoning more effectively handles intricate logical operations such as counting and numerical comparisons.
On GQA, out method outperforms previous zero-shot state-of-the-art, PNP-VQA, by $0.16$ ($0.42$ to $0.58$).
For object and face tagging, without training or fine-tuning, our method achieves 67.61\% and 60.82\% semantic correctness rates (Table \ref{tab:object-tagging-and-editing-performance}).

\begin{figure}[tb!]
    \centering
    \footnotesize
    \includegraphics[width=0.9\linewidth]{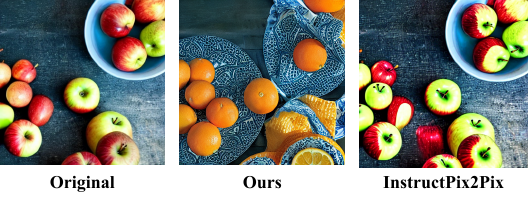}
    \vspace{-10px}
    \begin{lstlisting}[frame=single,xrightmargin=5px]
Instruction: Replace the bowl with something
else, and change the apples to other fruits.
    \end{lstlisting}
    \caption{
        Qualitative comparison of image editing.
        Compared to InstructPix2Pix, our image editing method follows the instructed edits better, as it successfully changed the bowl into plate and apples to oranges.
    }
    \label{fig:qualitative}
\end{figure}

\paragraph{Image generation and editing.}

For image generation and editing, we apply our technique to the OFCP and IGP20 datasets.
We rely on manual inspection for evaluating our performance on the OFCP dataset, and we observe 37 correctly edited images out of the 50 evaluated ones, resulting in a 74\% semantic correctness rate (Table \ref{tab:object-tagging-and-editing-performance}).
For IGP20, we choose as the baseline a diffusion model, InstructPix2Pix \cite{brooks2023instructpix2pix}, which also combines GPT-3 with image editing.
We show one example baseline comparison illustrated in Figure \ref{fig:qualitative}.

\begin{table}[t]
\centering
\footnotesize
  \setlength{\tabcolsep}{4pt}
  \begin{tabular}{l  
  >{\centering\arraybackslash}m{2cm}
  >{\centering\arraybackslash}m{2cm}
  >{\centering\arraybackslash}m{2cm}}
    \toprule
    \multirow{2}{*}{\textbf{Method}} &
    \multicolumn{2}{c}{\textbf{Visual Object Tagging}} & 
    \textbf{Image Editing} 
    \\ \cmidrule{2-4}
    &
    VQAR &
    OFCP &
    OFCP
    \\ \midrule
    Ours & 67.61\% & 60.82\% & 74.00\% \\ 
    \bottomrule
  \end{tabular}
  \vspace{-5px}
\captionof{table}{
Quantitative results on object tagging and image editing tasks.
We manually evaluate the tagged entities and the edited images for semantic correctness rates.
}
\label{tab:object-tagging-and-editing-performance}
\vspace{-2px}
\end{table}







\section{Conclusion}
\label{sec:conclusions}


We introduced \tool{}, a declarative framework designed for relational programming with foundation models. 
\tool{} brings together foundation models from diverse domains, providing a unified interface for composition and the ability to perform probabilistic logical reasoning. 
This results in solutions with comparable and often superior performance than neural-based baselines.
In the future, we aim to extend the capabilities of \tool{} beyond the current in-context learning settings to weakly-supervised training and fine-tuning of foundation models in an end-to-end manner. 

\section*{Acknowledgements}

We thank the anonymous reviewers for useful feedback.
This research was supported by NSF grant \#2313010 and DARPA grant \#FA8750-23-C-0080.
Ziyang Li was supported by an Amazon Fellowship.

\bibliography{references}

\begin{thebibliography}{55}
\providecommand{\natexlab}[1]{#1}

\bibitem[{Abiteboul, Hull, and Vianu(1994)}]{alice}
Abiteboul, S.; Hull, R.; and Vianu, V. 1994.
\newblock \emph{Foundations of Databases: {T}he Logical Level}.
\newblock Pearson, 1st edition.

\bibitem[{Adadi(2021)}]{adadi2021survey}
Adadi, A. 2021.
\newblock A survey on data-efficient algorithms in big data era.
\newblock \emph{Journal of Big Data}, 8(1): 24.

\bibitem[{Beurer-Kellner, Fischer, and Vechev(2022)}]{beurer2022prompting}
Beurer-Kellner, L.; Fischer, M.; and Vechev, M. 2022.
\newblock Prompting Is Programming: A Query Language For Large Language Models.
\newblock In \emph{PLDI}.

\bibitem[{Bommasani et~al.(2021)Bommasani, Hudson, Adeli, Altman, Arora, von Arx, Bernstein, Bohg, Bosselut, Brunskill, and et~al.}]{rishi2021foundationmodels}
Bommasani, R.; Hudson, D.~A.; Adeli, E.; Altman, R.~B.; Arora, S.; von Arx, S.; Bernstein, M.~S.; Bohg, J.; Bosselut, A.; Brunskill, E.; and et~al. 2021.
\newblock On the Opportunities and Risks of Foundation Models.
\newblock arXiv:2108.07258.

\bibitem[{Brooks, Holynski, and Efros(2023)}]{brooks2023instructpix2pix}
Brooks, T.; Holynski, A.; and Efros, A.~A. 2023.
\newblock InstructPix2Pix: Learning to Follow Image Editing Instructions.
\newblock arXiv:2211.09800.

\bibitem[{Bubeck et~al.(2023)Bubeck, Chandrasekaran, Eldan, Gehrke, Horvitz, Kamar, Lee, Lee, Li, Lundberg et~al.}]{bubeck2023sparks}
Bubeck, S.; Chandrasekaran, V.; Eldan, R.; Gehrke, J.; Horvitz, E.; Kamar, E.; Lee, P.; Lee, Y.~T.; Li, Y.; Lundberg, S.; et~al. 2023.
\newblock Sparks of artificial general intelligence: Early experiments with GPT-4.
\newblock arXiv:2303.12712.

\bibitem[{Chen et~al.(2020)Chen, Liang, Yu, Zhou, Song, and Le}]{chen2020nerd}
Chen, X.; Liang, C.; Yu, A.~W.; Zhou, D.; Song, D.; and Le, Q.~V. 2020.
\newblock Neural Symbolic Reader: Scalable Integration of Distributed and Symbolic Representations for Reading Comprehension.
\newblock In \emph{ICLR}.

\bibitem[{Cobbe et~al.(2021)Cobbe, Kosaraju, Bavarian, Chen, Jun, Kaiser, Plappert, Tworek, Hilton, Nakano, Hesse, and Schulman}]{cobbe2021training}
Cobbe, K.; Kosaraju, V.; Bavarian, M.; Chen, M.; Jun, H.; Kaiser, L.; Plappert, M.; Tworek, J.; Hilton, J.; Nakano, R.; Hesse, C.; and Schulman, J. 2021.
\newblock Training Verifiers to Solve Math Word Problems.
\newblock arXiv:2110.14168.

\bibitem[{Davis and Aaronson(2023)}]{davis2023testing}
Davis, E.; and Aaronson, S. 2023.
\newblock Testing GPT-4 with Wolfram Alpha and Code Interpreter plug-ins on math and science problems.
\newblock arXiv:2308.05713.

\bibitem[{Gao et~al.(2023)Gao, Madaan, Zhou, Alon, Liu, Yang, Callan, and Neubig}]{gao2023pal}
Gao, L.; Madaan, A.; Zhou, S.; Alon, U.; Liu, P.; Yang, Y.; Callan, J.; and Neubig, G. 2023.
\newblock PAL: Program-aided Language Models.
\newblock arXiv:2211.10435.

\bibitem[{Gupta and Kembhavi(2022)}]{Gupta2022VisProg}
Gupta, T.; and Kembhavi, A. 2022.
\newblock Visual Programming: Compositional visual reasoning without training.
\newblock arXiv:2211.11559.

\bibitem[{Hertz et~al.(2022)Hertz, Mokady, Tenenbaum, Aberman, Pritch, and Cohen-Or}]{hertz2022prompttoprompt}
Hertz, A.; Mokady, R.; Tenenbaum, J.; Aberman, K.; Pritch, Y.; and Cohen-Or, D. 2022.
\newblock Prompt-to-Prompt Image Editing with Cross Attention Control.
\newblock arXiv:2208.01626.

\bibitem[{Huang et~al.(2021)Huang, Li, Chen, Samel, Naik, Song, and Si}]{huang2021scallop}
Huang, J.; Li, Z.; Chen, B.; Samel, K.; Naik, M.; Song, L.; and Si, X. 2021.
\newblock Scallop: From Probabilistic Deductive Databases to Scalable Differentiable Reasoning.
\newblock In \emph{NeurIPS}.

\bibitem[{Hudson and Manning(2019)}]{drew2019gqa}
Hudson, D.~A.; and Manning, C.~D. 2019.
\newblock {GQA:} a new dataset for compositional question answering over real-world images.
\newblock arXiv:1902.09506.

\bibitem[{Johnson et~al.(2016)Johnson, Hariharan, van~der Maaten, Fei{-}Fei, Zitnick, and Girshick}]{johnson2016clevr}
Johnson, J.; Hariharan, B.; van~der Maaten, L.; Fei{-}Fei, L.; Zitnick, C.~L.; and Girshick, R.~B. 2016.
\newblock {CLEVR:} {A} Diagnostic Dataset for Compositional Language and Elementary Visual Reasoning.
\newblock arXiv:1612.06890.

\bibitem[{Kim, Son, and Kim(2021)}]{kim2021vilt}
Kim, W.; Son, B.; and Kim, I. 2021.
\newblock ViLT: Vision-and-Language Transformer Without Convolution or Region Supervision.
\newblock arXiv:2102.03334.

\bibitem[{Kirillov et~al.(2023)Kirillov, Mintun, Ravi, Mao, Rolland, Gustafson, Xiao, Whitehead, Berg, Lo et~al.}]{kirillov2023segment}
Kirillov, A.; Mintun, E.; Ravi, N.; Mao, H.; Rolland, C.; Gustafson, L.; Xiao, T.; Whitehead, S.; Berg, A.~C.; Lo, W.-Y.; et~al. 2023.
\newblock Segment Anything.
\newblock arXiv:2304.02643.

\bibitem[{Kojima et~al.(2022)Kojima, Gu, Reid, Matsuo, and Iwasawa}]{kojima2022large}
Kojima, T.; Gu, S.~S.; Reid, M.; Matsuo, Y.; and Iwasawa, Y. 2022.
\newblock Large language models are zero-shot reasoners.
\newblock In \emph{NeurIPS}.

\bibitem[{Li et~al.(2018)Li, Wang, Wang, Tai, Qian, Yang, Wang, Li, and Huang}]{jian2018dsfd}
Li, J.; Wang, Y.; Wang, C.; Tai, Y.; Qian, J.; Yang, J.; Wang, C.; Li, J.; and Huang, F. 2018.
\newblock {DSFD:} Dual Shot Face Detector.
\newblock arXiv:1810.10220.

\bibitem[{Li et~al.(2020)Li, Huang, Hong, Chen, Wu, and Zhu}]{li2020closed}
Li, Q.; Huang, S.; Hong, Y.; Chen, Y.; Wu, Y.~N.; and Zhu, S.-C. 2020.
\newblock Closed Loop Neural-Symbolic Learning via Integrating Neural Perception, Grammar Parsing, and Symbolic Reasoning.
\newblock In \emph{ICML}.

\bibitem[{Li, Lei, and Yang(2022)}]{li2022fe2h}
Li, X.-Y.; Lei, W.-J.; and Yang, Y.-B. 2022.
\newblock From Easy to Hard: Two-stage Selector and Reader for Multi-hop Question Answering.
\newblock arXiv:2205.11729.

\bibitem[{Li, Huang, and Naik(2023)}]{li2023scallop}
Li, Z.; Huang, J.; and Naik, M. 2023.
\newblock Scallop: A Language for Neurosymbolic Programming.
\newblock In \emph{PLDI}.

\bibitem[{Liang et~al.(2023)Liang, Wu, Song, Wu, Xia, Liu, Ou, Lu, Ji, Mao, Wang, Shou, Gong, and Duan}]{liang2023taskmatrixai}
Liang, Y.; Wu, C.; Song, T.; Wu, W.; Xia, Y.; Liu, Y.; Ou, Y.; Lu, S.; Ji, L.; Mao, S.; Wang, Y.; Shou, L.; Gong, M.; and Duan, N. 2023.
\newblock TaskMatrix.AI: Completing Tasks by Connecting Foundation Models with Millions of APIs.
\newblock arXiv:2303.16434.

\bibitem[{Liu et~al.(2019)Liu, Ott, Goyal, Du, Joshi, Chen, Levy, Lewis, Zettlemoyer, and Stoyanov}]{liu2019roberta}
Liu, Y.; Ott, M.; Goyal, N.; Du, J.; Joshi, M.; Chen, D.; Levy, O.; Lewis, M.; Zettlemoyer, L.; and Stoyanov, V. 2019.
\newblock RoBERTa: {A} Robustly Optimized {BERT} Pretraining Approach.
\newblock arXiv:1907.11692.

\bibitem[{Lyu et~al.(2023)Lyu, Havaldar, Stein, Zhang, Rao, Wong, Apidianaki, and Callison-Burch}]{lyu2023faithful}
Lyu, Q.; Havaldar, S.; Stein, A.; Zhang, L.; Rao, D.; Wong, E.; Apidianaki, M.; and Callison-Burch, C. 2023.
\newblock Faithful Chain-of-Thought Reasoning.
\newblock arXiv:2301.13379.

\bibitem[{Manhaeve et~al.(2018)Manhaeve, Dumancic, Kimmig, Demeester, and Raedt}]{robin2018deepproblog}
Manhaeve, R.; Dumancic, S.; Kimmig, A.; Demeester, T.; and Raedt, L.~D. 2018.
\newblock DeepProbLog: Neural Probabilistic Logic Programming.
\newblock arXiv:1805.10872.

\bibitem[{Mao et~al.(2019)Mao, Gan, Kohli, Tenenbaum, and Wu}]{mao2019nscl}
Mao, J.; Gan, C.; Kohli, P.; Tenenbaum, J.~B.; and Wu, J. 2019.
\newblock The Neuro-Symbolic Concept Learner: Interpreting Scenes, Words, and Sentences From Natural Supervision.
\newblock arXiv:1904.12584.

\bibitem[{McKenna et~al.(2023)McKenna, Li, Cheng, Hosseini, Johnson, and Steedman}]{mckenna2023sources}
McKenna, N.; Li, T.; Cheng, L.; Hosseini, M.~J.; Johnson, M.; and Steedman, M. 2023.
\newblock Sources of Hallucination by Large Language Models on Inference Tasks.
\newblock arXiv:2305.14552.

\bibitem[{Minderer et~al.(2022)Minderer, Gritsenko, Stone, Neumann, Weissenborn, Dosovitskiy, Mahendran, Arnab, Dehghani, Shen, Wang, Zhai, Kipf, and Houlsby}]{minderer2022simple}
Minderer, M.; Gritsenko, A.; Stone, A.; Neumann, M.; Weissenborn, D.; Dosovitskiy, A.; Mahendran, A.; Arnab, A.; Dehghani, M.; Shen, Z.; Wang, X.; Zhai, X.; Kipf, T.; and Houlsby, N. 2022.
\newblock Simple Open-Vocabulary Object Detection with Vision Transformers.
\newblock arXiv:2205.06230.

\bibitem[{Minervini et~al.(2020)Minervini, Riedel, Stenetorp, Grefenstette, and Rockt{\"a}schel}]{minervini2020ctp}
Minervini, P.; Riedel, S.; Stenetorp, P.; Grefenstette, E.; and Rockt{\"a}schel, T. 2020.
\newblock Learning Reasoning Strategies in End-to-End Differentiable Proving.
\newblock In \emph{ICML}.

\bibitem[{Nakano et~al.(2021)Nakano, Hilton, Balaji, Wu, Ouyang, Kim, Hesse, Jain, Kosaraju, Saunders, Jiang, Cobbe, Eloundou, Krueger, Button, Knight, Chess, and Schulman}]{reiichiro2021webgpt}
Nakano, R.; Hilton, J.; Balaji, S.; Wu, J.; Ouyang, L.; Kim, C.; Hesse, C.; Jain, S.; Kosaraju, V.; Saunders, W.; Jiang, X.; Cobbe, K.; Eloundou, T.; Krueger, G.; Button, K.; Knight, M.; Chess, B.; and Schulman, J. 2021.
\newblock WebGPT: Browser-assisted question-answering with human feedback.
\newblock arXiv:2112.09332.

\bibitem[{Nogueira and Cho(2019)}]{nogueira2019passage}
Nogueira, R.; and Cho, K. 2019.
\newblock Passage Re-ranking with BERT.
\newblock arXiv:1901.04085.

\bibitem[{OpenAI(2023)}]{openai2023gpt4}
OpenAI. 2023.
\newblock GPT-4 Technical Report.
\newblock arXiv:2303.08774.

\bibitem[{Radford et~al.(2021)Radford, Kim, Hallacy, Ramesh, Goh, Agarwal, Sastry, Askell, Mishkin, Clark, Krueger, and Sutskever}]{alec2021transferablevisualmodels}
Radford, A.; Kim, J.~W.; Hallacy, C.; Ramesh, A.; Goh, G.; Agarwal, S.; Sastry, G.; Askell, A.; Mishkin, P.; Clark, J.; Krueger, G.; and Sutskever, I. 2021.
\newblock Learning Transferable Visual Models From Natural Language Supervision.
\newblock arXiv:2103.00020.

\bibitem[{Rajasekharan et~al.(2023)Rajasekharan, Zeng, Padalkar, and Gupta}]{rajasekharan2023star}
Rajasekharan, A.; Zeng, Y.; Padalkar, P.; and Gupta, G. 2023.
\newblock Reliable Natural Language Understanding with Large Language Models and Answer Set Programming.
\newblock In \emph{International Conference on Logic Programming}.

\bibitem[{Ramesh et~al.(2021)Ramesh, Pavlov, Goh, Gray, Voss, Radford, Chen, and Sutskever}]{ramesh2021zero}
Ramesh, A.; Pavlov, M.; Goh, G.; Gray, S.; Voss, C.; Radford, A.; Chen, M.; and Sutskever, I. 2021.
\newblock Zero-shot text-to-image generation.
\newblock In \emph{ICML}.

\bibitem[{Ratner et~al.(2023)Ratner, Levine, Belinkov, Ram, Magar, Abend, Karpas, Shashua, Leyton-Brown, and Shoham}]{ratner2023parallel}
Ratner, N.; Levine, Y.; Belinkov, Y.; Ram, O.; Magar, I.; Abend, O.; Karpas, E.; Shashua, A.; Leyton-Brown, K.; and Shoham, Y. 2023.
\newblock Parallel Context Windows for Large Language Models.
\newblock In \emph{Proceedings of the ACL}.

\bibitem[{Reddy et~al.(2022)Reddy, Màrquez, Valero, Rao, Zaragoza, Bandyopadhyay, Biswas, Xing, and Subbian}]{reddy2022shopping}
Reddy, C.~K.; Màrquez, L.; Valero, F.; Rao, N.; Zaragoza, H.; Bandyopadhyay, S.; Biswas, A.; Xing, A.; and Subbian, K. 2022.
\newblock Shopping Queries Dataset: A Large-Scale ESCI Benchmark for Improving Product Search.
\newblock arXiv:2206.06588.

\bibitem[{Richards(2023)}]{autogpt}
Richards, T.~B. 2023.
\newblock {AutoGPT}.
\newblock \url{https://github.com/Significant-Gravitas/AutoGPT}.
\newblock Accessed: 2024-02-12.

\bibitem[{Rombach et~al.(2022)Rombach, Blattmann, Lorenz, Esser, and Ommer}]{Rombach_2022_diffusion}
Rombach, R.; Blattmann, A.; Lorenz, D.; Esser, P.; and Ommer, B. 2022.
\newblock High-Resolution Image Synthesis With Latent Diffusion Models.
\newblock In \emph{CVPR}.

\bibitem[{Schick et~al.(2023)Schick, Dwivedi-Yu, Dessì, Raileanu, Lomeli, Zettlemoyer, Cancedda, and Scialom}]{schick2023toolformer}
Schick, T.; Dwivedi-Yu, J.; Dessì, R.; Raileanu, R.; Lomeli, M.; Zettlemoyer, L.; Cancedda, N.; and Scialom, T. 2023.
\newblock Toolformer: Language Models Can Teach Themselves to Use Tools.
\newblock arXiv:2302.04761.

\bibitem[{Sinha et~al.(2019)Sinha, Sodhani, Dong, Pineau, and Hamilton}]{sinha2019clutrr}
Sinha, K.; Sodhani, S.; Dong, J.; Pineau, J.; and Hamilton, W.~L. 2019.
\newblock CLUTRR: A Diagnostic Benchmark for Inductive Reasoning from Text.
\newblock arXiv:1908.06177.

\bibitem[{Song et~al.(2020)Song, Tan, Qin, Lu, and Liu}]{song2020mpnet}
Song, K.; Tan, X.; Qin, T.; Lu, J.; and Liu, T.-Y. 2020.
\newblock MPNet: Masked and Permuted Pre-training for Language Understanding.
\newblock arXiv:2004.09297.

\bibitem[{Srivastava et~al.(2023)Srivastava, Rastogi, Rao, Shoeb, Abid, Fisch, Brown, Santoro, Gupta, Garriga-Alonso, and et~al.}]{srivastava2023imitation}
Srivastava, A.; Rastogi, A.; Rao, A.; Shoeb, A. A.~M.; Abid, A.; Fisch, A.; Brown, A.~R.; Santoro, A.; Gupta, A.; Garriga-Alonso, A.; and et~al. 2023.
\newblock Beyond the Imitation Game: Quantifying and extrapolating the capabilities of language models.
\newblock arXiv:2206.04615.

\bibitem[{Tiong et~al.(2022)Tiong, Li, Li, Savarese, and Hoi}]{tiong-2022-pnpvqa}
Tiong, A. M.~H.; Li, J.; Li, B.; Savarese, S.; and Hoi, S.~C. 2022.
\newblock Plug-and-Play {VQA}: Zero-shot {VQA} by Conjoining Large Pretrained Models with Zero Training.
\newblock In Goldberg, Y.; Kozareva, Z.; and Zhang, Y., eds., \emph{Findings of the ACL: EMNLP}.

\bibitem[{Touvron et~al.(2023)Touvron, Martin, Stone, Albert, Almahairi, Babaei, Bashlykov, Batra, Bhargava, Bhosale et~al.}]{touvron2023llama}
Touvron, H.; Martin, L.; Stone, K.; Albert, P.; Almahairi, A.; Babaei, Y.; Bashlykov, N.; Batra, S.; Bhargava, P.; Bhosale, S.; et~al. 2023.
\newblock Llama 2: Open Foundation and Fine-Tuned Chat Models.
\newblock arXiv:2307.09288.

\bibitem[{Wang et~al.(2019)Wang, Donti, Wilder, and Kolter}]{wang2019satnet}
Wang, P.-W.; Donti, P.~L.; Wilder, B.; and Kolter, Z. 2019.
\newblock SATNet: Bridging Deep Learning and Logical Reasoning Using a Differentiable Satisfiability Solver.
\newblock In \emph{ICML}.

\bibitem[{Wang et~al.(2023)Wang, Wei, Schuurmans, Le, Chi, Narang, Chowdhery, and Zhou}]{wang2023selfconsistency}
Wang, X.; Wei, J.; Schuurmans, D.; Le, Q.; Chi, E.; Narang, S.; Chowdhery, A.; and Zhou, D. 2023.
\newblock Self-Consistency Improves Chain of Thought Reasoning in Language Models.
\newblock arXiv:2203.11171.

\bibitem[{Wei et~al.(2023)Wei, Wang, Schuurmans, Bosma, Ichter, Xia, Chi, Le, and Zhou}]{wei2023chainofthought}
Wei, J.; Wang, X.; Schuurmans, D.; Bosma, M.; Ichter, B.; Xia, F.; Chi, E.; Le, Q.; and Zhou, D. 2023.
\newblock Chain-of-Thought Prompting Elicits Reasoning in Large Language Models.
\newblock arXiv:2201.11903.

\bibitem[{Xu et~al.(2022)Xu, Rawat, Wong, Kankanhalli, and Shah}]{xu2022dont}
Xu, Z.; Rawat, Y.~S.; Wong, Y.; Kankanhalli, M.; and Shah, M. 2022.
\newblock Don't Pour Cereal into Coffee: Differentiable Temporal Logic for Temporal Action Segmentation.
\newblock In \emph{NeurIPS}.

\bibitem[{Yang et~al.(2018)Yang, Qi, Zhang, Bengio, Cohen, Salakhutdinov, and Manning}]{yang2018hotpotqa}
Yang, Z.; Qi, P.; Zhang, S.; Bengio, Y.; Cohen, W.~W.; Salakhutdinov, R.; and Manning, C.~D. 2018.
\newblock HotpotQA: A dataset for diverse, explainable multi-hop question answering.
\newblock arXiv:1809.09600.

\bibitem[{Yao et~al.(2023)Yao, Zhao, Yu, Du, Shafran, Narasimhan, and Cao}]{yao2023react}
Yao, S.; Zhao, J.; Yu, D.; Du, N.; Shafran, I.; Narasimhan, K.; and Cao, Y. 2023.
\newblock ReAct: Synergizing Reasoning and Acting in Language Models.
\newblock arXiv:2210.03629.

\bibitem[{Yi et~al.(2018)Yi, Wu, Gan, Torralba, Kohli, and Tenenbaum}]{yi2018nsvqa}
Yi, K.; Wu, J.; Gan, C.; Torralba, A.; Kohli, P.; and Tenenbaum, J. 2018.
\newblock Neural-Symbolic VQA: Disentangling Reasoning from Vision and Language Understanding.
\newblock In \emph{NeurIPS}.

\bibitem[{Yin et~al.(2022)Yin, Wang, Wu, Yan, Hu, Zhang, Cao, Huang, and Qiu}]{yin2022c2fm}
Yin, Z.; Wang, Y.; Wu, Y.; Yan, H.; Hu, X.; Zhang, X.; Cao, Z.; Huang, X.; and Qiu, X. 2022.
\newblock Rethinking Label Smoothing on Multi-hop Question Answering.
\newblock arXiv:2212.09512.

\bibitem[{Zheng et~al.(2023)Zheng, Chiang, Sheng, Zhuang, Wu, Zhuang, Lin, Li, Li, Xing, Zhang, Gonzalez, and Stoica}]{zheng2023vicuna}
Zheng, L.; Chiang, W.-L.; Sheng, Y.; Zhuang, S.; Wu, Z.; Zhuang, Y.; Lin, Z.; Li, Z.; Li, D.; Xing, E.~P.; Zhang, H.; Gonzalez, J.~E.; and Stoica, I. 2023.
\newblock Judging LLM-as-a-judge with MT-Bench and Chatbot Arena.
\newblock arXiv:2306.05685.

\end{thebibliography}

\newpage
\appendix
\section{Full \tool{} Language}
\label{app:full_language}

We present the full surface syntax of the \tool{} language in Fig. \ref{fig:surface-syntax}.

\begin{figure*}
\begin{lstlisting}
ITEM ::= ATTRS? DECL
ATTRS ::= @ATTR | @ATTR(POS_ARG, ..., KW=KW_ARG, ...)
DECL ::= IMPORT_DECL | TYPE_DECL | CONST_DECL | REL_DECL | QUERY_DECL
IMPORT_DECL ::= import "FILENAME"
TYPE_DECL ::= type TYPE_ALIAS = TYPE | type SUB_TYPE <: TYPE 
            | (type TYPE = CONS1(TYPE, ...) | CONS2(TYPE, ...) | ...)
            | type PRED(ARG: TYPE, ...) | type $FN(ARG: TYPE, ...) -> TYPE
CONST_DECL ::= const NAME = CONSTANT
REL_DECL ::= rel [PROB::]PRED(EXPR, ...) | rel PRED = {[PROB::](EXPR, ...), ...} 
           | rel [PROB::]ATOM = FORMULA | rel { ATOM; ... } = FORMULA
QUERY_DECL ::= query PRED
TYPE ::= i8 | i16 | i32 | i64 | isize | u8 | u16 | u32 | u64 | usize | f32 | f64 | char | bool | String 
       | DateTime | Duration | Tensor | NAME
CONSTANT ::= NUMBER | BOOL_LITERAL | CHAR_LITERAL | STRING_LITERAL | DATETIME_LITERAL | DURATION_LITERAL
EXPR ::= CONSTANT | EXPR BIN_OP EXPR | UNA_OP EXPR | if EXPR then EXPR else EXPR | EXPR as TYPE 
       | $FN(EXPR, ...) | new CONS(EXPR, ...)
FORMULA ::= PRED(EXPR, ...) | FORMULA and FORMULA | FORMULA or FORMULA | not FORMULA
          | case VAR is ENTITY | VAR := AGGREGATOR(VAR*: FORMULA [where VAR*: FORMULA])
ENTITY ::= EXPR | CONS(ENTITY, ...)
\end{lstlisting}
\caption{Surface Syntax of \tool{} language.}
\label{fig:surface-syntax}
\end{figure*}

\section{Detailed Example}
\label{app:detailed_example}

In this section, we describe the \tool{} program for one of our benchmark applications, CLEVR \cite{johnson2016clevr}.
We decompose this application into three sub-tasks:
1) extracting a structured scene graph from the input image,
2) extracting an executable query program from the input natural language (NL) question, and
3) combining both to answer the question based on the scene graph.
We next describe how we solve each of these sub-tasks.
For illustration, we use the example image and question shown in Fig. \ref{fig:clevr-example}.

\subsection{Image to structured scene graph}

To convert image to structured scene graph, we use two vision models, namely OWL-ViT \cite{minderer2022simple} and CLIP \cite{alec2021transferablevisualmodels}.
We use OWL-ViT for obtaining object segments and CLIP models for classifying object properties.
The goal is to construct scene graph which contains the following information: the shape, color, material, and size for each object, and the spatial relationships between each pair of objects.

Our object detection predicate is defined as follows:
\begin{lstlisting}[language=scallop]
@owl_vit(["cube", "sphere", "cylinder"],
  expand_crop_region=10, limit=10,
  flatten_probability=true)
type segment_image(
  bound img: Tensor, id: u32, 
  cropped_image: Tensor, area: u32, 
  bbox_center_x: u32, bbox_bottom_y: u32)
\end{lstlisting}
We are using the \inlinecode{@owl\_vit} foreign attribute to decorate a predicate \inlinecode{vit\_segment\_image}.
Here, the image has one bounded argument which is the input image, and it produces image segments represented by 5 tuples, containing segment id (\inlinecode{id}), segmented image (\inlinecode{cropped\_image}), the area of segment (\inlinecode{area}), the center $x$ coordinate (\inlinecode{bbox\_center\_x}), and the bottom $y$ coordinate (\inlinecode{bbox\_bottom\_y}).
Specifically, segmented images can be passed to downstream image classifiers, the area is used to classify whether the object is big or small, and the coordinates are used to determine spatial relationships between objects.

Note that the arguments we pass to \inlinecode{@owl\_vit} contain expected labels of \inlinecode{cube}, \inlinecode{sphere}, and \inlinecode{cylinder}.
Because OWL-ViT does not perform well at classifying given geometric objects by shape, we do not use it to query the labels associated with each object.
Rather, these labels identify the image segments the model extracts from the base image.

We set \inlinecode{expand\_crop\_region} to be 10, which expands the cropped images by the given factor.
Since the bounding boxes of the objects are tight, enlarging the crop region can help subsequent classifiers to better see the object.
With the \inlinecode{limit} set to 10, OWL-ViT only generates 10 image segments.
Lastly, we set \inlinecode{flatten\_probability} to be \inlinecode{true}.
Again, OWL-ViT is not trained on CLEVR, so it produces very low confidence scores on all recognized objects.
In order to not let the scores affect downstream computation, we overwrite the probability to 1 for all objects.

We load the image specified by the image directory path using the foreign function \inlinecode{\$load\_image}, and then segment the image using the \inlinecode{segment\_image} predicate:
\begin{lstlisting}[language=scallop]
type img_dir(directory: String) // input
rel image($load_image(d)) = img_dir(d) // load
rel obj_seg(id, seg, obj_size, x, y) = 
  image(img) and 
  segment_image(img, id, seg, obj_size, x, y)
rel obj(id) = obj_seg(id, _, _, _, _)
\end{lstlisting}

We next define the shape classifier. For this, we repurpose \inlinecode{@clip} to classify each object segment with a label from three possible shapes: \inlinecode{cube}, \inlinecode{sphere}, and \inlinecode{cylinder}.
In order to interface with CLIP, we create a prompt \inlinescl{"a \{\{\}\} shaped object"}.
Each label is filled into the prompt, producing short phrases like ``a cube shaped object''.
Then, the three prompts are passed to CLIP along with the object image, and facts with labels are returned with probabilities.
\begin{lstlisting}[language=scallop]
@clip(["cube", "cylinder", "sphere"], 
  prompt="a {{}} shaped object")
type classify_shape(
  bound obj_img: Tensor, 
  shape: String)
rel shape(o, s) = obj_seg(o, seg, _, _, _) and 
  classify_shape(seg, s)
\end{lstlisting}
The classifiers for color and material are done similarly:
\begin{lstlisting}[language=scallop]
@clip(["red", "blue", "yellow", 
  "purple", "gray", "brown", "cyan", "green"], 
  prompt="a {{}} colored object")
type classify_color(
  bound obj_img: Tensor, 
  color: String)
rel color(o, c) = obj_seg(o, seg, _, _, _) and 
  classify_color(seg, c)
\end{lstlisting}
\begin{lstlisting}[language=scallop]
@clip(["shiny metal", "solid rubber"], 
  prompt="object made out of {{}} material")
type classify_material(
  bound obj_img: Tensor, 
  material: String)
rel material(o, m) = obj_seg(o, s, _, _, _) and
  classify_material(s, m)
\end{lstlisting}
From here, we just invoke the previous 
We continue to discuss how do we obtain the size and spatial relationships.
In order to obtain the size (\inlinecode{small} or \inlinecode{large}) of each object, we use a probabilistic rule for specifying that:
\begin{lstlisting}[language=scallop]
rel {
  0.9::size(o, if l then "large" else "small"); 
  0.1::size(o, if !l then "large" else "small")
} = obj_seg(o, _, size, _, _) and l == size > A
\end{lstlisting}
Finally, the spatial relationship (\inlinecode{left}, \inlinecode{right}, \inlinecode{front}, and \inlinecode{behind}) is derived from object coordinates.
\begin{lstlisting}[language=scallop]
rel obj_pos(o, x, y) = obj_seg(o, _, _, x, y)
rel relate(o1, o2, dir) = o1 != o2 and
  obj_pos(o1, x1, _) and obj_pos(o2, x2, _) and
  dir == if x1 < x2 then "left" else "right"
rel relate(o1, o2, dir) = o1 != o2 and
  obj_pos(o1, _, y1) and obj_pos(o2, _, y2) and 
  dir == if y1 > y2 then "front" else "behind"
\end{lstlisting}
Combining everything together, we have produced the relationships
\inlinecode{color}, \inlinecode{shape}, \inlinecode{material}, \inlinecode{size}, and \inlinecode{relate}, forming the scene graph of the image.

\subsection{NL question to programmatic query}

We use the GPT-4 model \cite{openai2023gpt4} for converting a natural language question into a programmatic query.
The first step is defining the domain specific language (DSL) for querying the CLEVR dataset:
\begin{lstlisting}[language=scallop]
type Query = Scene()
           | FilterShape(Query, String)
           | FilterMaterial(Query, String)
           | FilterColor(Query, String)
           | FilterSize(Query, String)
           | Relate(Query, String)
           | Count(Query)
           | Exists(Query)
           | GreaterThan(Query, Query)
           | LessThan(Query, Query)
           | Equals(Query, Query)
           | Intersect(Query, Query)
           | Union(Query, Query)
           | SameSize(Query)
           | SameColor(Query)
           | SameMaterial(Query)
           | SameShape(Query)
           | QueryMaterial(Query)
           | QueryColor(Query)
           | QueryShape(Query)
           | QuerySize(Query)
\end{lstlisting}
Notice that the DSL is represented by the user-defined algebraic data type (ADT) \inlinecode{Query}, which contains constructs for getting objects, counting objects, checking existence of objects, and even comparing counts obtained from evaluating multiple queries.
We then create the semantic parser for the DSL by configuring the GPT-4 model:
\begin{lstlisting}[language=scallop]
@gpt_semantic_parse(
  header="""
Please convert a question into its programmatic 
form according to the following language:
Expr = Scene() | FilterShape(Expr, String) | ...

Please pick shapes among \"cylinder\", ...;
Colors are among \"red\", \"blue\", ...;
Materials are among \"shiny metal\" and ...;
Sizes are among \"large\" and \"small\";
Spatial relations are among \"left\", ...""",
  prompt="""
Question: {{s}}
Query: {{e}}""",
  examples=[
    ("How many red objects are there?", 
     "Count(FilterColor(Scene(), \"red\"))"),
    ("Is there a cube?", 
     "Exists(FilterShape(Scene(), \"cube\"))"),
    ...],
  model="gpt-4")
type parse_query(bound s: String, q: Query)
\end{lstlisting}
Other than the \inlinecode{model} argument which is used to specify the OpenAI model to call, we also pass 3 additional arguments to \inlinecode{gpt\_semantic\_parse}: \inlinecode{header}, \inlinecode{prompt}, and \inlinecode{examples}.
These arguments construct the first part of the prompt that we pass to GPT-4.
Assuming the actual question (``Is there an object to the left of the cube?'') is passed to the foreign predicate \inlinecode{parse\_expr} as the first argument \inlinescl{s}, the entire prompt becomes:
\begin{lstlisting}[frame=single]
# header part...
Please convert a question into its program...
# example 1
Question: How many red objects are there?
Query: Count(FilterColor(Scene(), "red"))
# example 2
Question: Is there a cube?
Query: Exists(FilterShape(Scene(), "cube"))
# more examples...
# actual question
Question: Is there an object to the left of 
the cube?
Query:

>>> GPT-4 Answer:
Exists(Relate(
    FilterShape(Scene(), "cube"), "left"))
\end{lstlisting}
Then, GPT-4 is prompted to produce the query, which is parsed back into our ADT \inlinecode{Query}:
\begin{lstlisting}[language=scallop]
type question(String) // input question string
rel parsed_query(q) = 
    question(s) and parse_query(s, q)
\end{lstlisting}

\subsection{Putting it all together}

The last part which brings everything together is the semantics of our \inlinecode{Query} DSL.
The semantics is inductively defined on the \inlinecode{Query} data structure.
We start from defining the variants which return a set of objects.
For this, we use the \inlinecode{eval\_obj} binary relation to connect queries with their evaluated object IDs:
\begin{lstlisting}[language=scallop]
rel eval_obj(e, o) =                    // Scene
    case e is Scene() and object(o)
rel eval_obj(e, o) =              // FilterShape
    case e is FilterShape(e1, s) and
    eval_obj(e1, o) and shape(o, s)
rel eval_obj(e, o) =              // FilterColor
    case e is FilterColor(e1, c) and
    eval_obj(e1, o) and color(o, c)
rel eval_obj(e, o) =           // FilterMaterial
    case e is FilterMaterial(e1, m) and
    eval_obj(e1, o) and material(o, m)
rel eval_obj(e, o) =               // FilterSize
    case e is FilterSize(e1, s) and
    eval_obj(e1, o) and size(o, s)
rel eval_obj(e, p) =                   // Relate
    case e is Relate(e1, dir) and
    eval_obj(e1, o) and relate(p, o, dir)
rel eval_obj(e, p) =                 // SameSize
    case e is SameSize(e1) and 
    eval_obj(e1, o) and size(o, s) and 
    size(p, s) and o != p
rel eval_obj(e, p) =                // SameColor
    case e is SameColor(e1) and 
    eval_obj(e1, o) and color(o, c) and 
    color(p, c) and o != p
rel eval_obj(e, p) =             // SameMaterial
    case e is SameMaterial(e1) and 
    eval_obj(e1, o) and material(o, m) and 
    material(p, m) and o != p
rel eval_obj(e, p) =                // SameShape
    case e is SameShape(e1) and 
    eval_obj(e1, o) and shape(o, s) and 
    shape(p, s) and o != p
\end{lstlisting}
We next define the semantics for queries which evaluate to a boolean, producing the \inlinescl{eval\_bool} relation:
\begin{lstlisting}[language=scallop]
rel eval_bool(e, b) = b := exists(     // Exists
    o: eval_obj(e1, o) where 
    e: case e is Exists(e1))
rel eval_bool(e, n1 > n2) =       // GreaterThan
    case e is GreaterThan(e1, e2) and
    eval_num(e1, n1) and eval_num(e2, n2)
rel eval_bool(e, n1 < n2) =          // LessThan
    case e is LessThan(e1, e2) and 
    eval_num(e1, n1) and eval_num(e2, n2)
rel eval_bool(e, n1 == n2) =           // Equals
    case e is Equals(e1, e2) and
    eval_num(e1, n1) and eval_num(e2, n2)
\end{lstlisting}
We finally define the semantics for queries which evaluate to a number, producing the \inlinescl{eval\_num} relation:
\begin{lstlisting}[language=scallop]
rel eval_num(e, n) = n := count(        // Count
    o: eval_obj(e1, o) where 
    e: case e is Count(e1))
\end{lstlisting}
To connect everything together, we apply the \inlinecode{eval\_*} relation on the parsed expression to get the evaluated result:
\begin{lstlisting}[language=scallop]
rel result(r as String) = 
    parsed_query(q) and eval_num(q, r)
rel result(r as String) = 
    parsed_query(q) and eval_bool(q, r)
\end{lstlisting}

\subsection{A concrete example}

\begin{figure}
\centering
\includegraphics[width=0.8\linewidth]{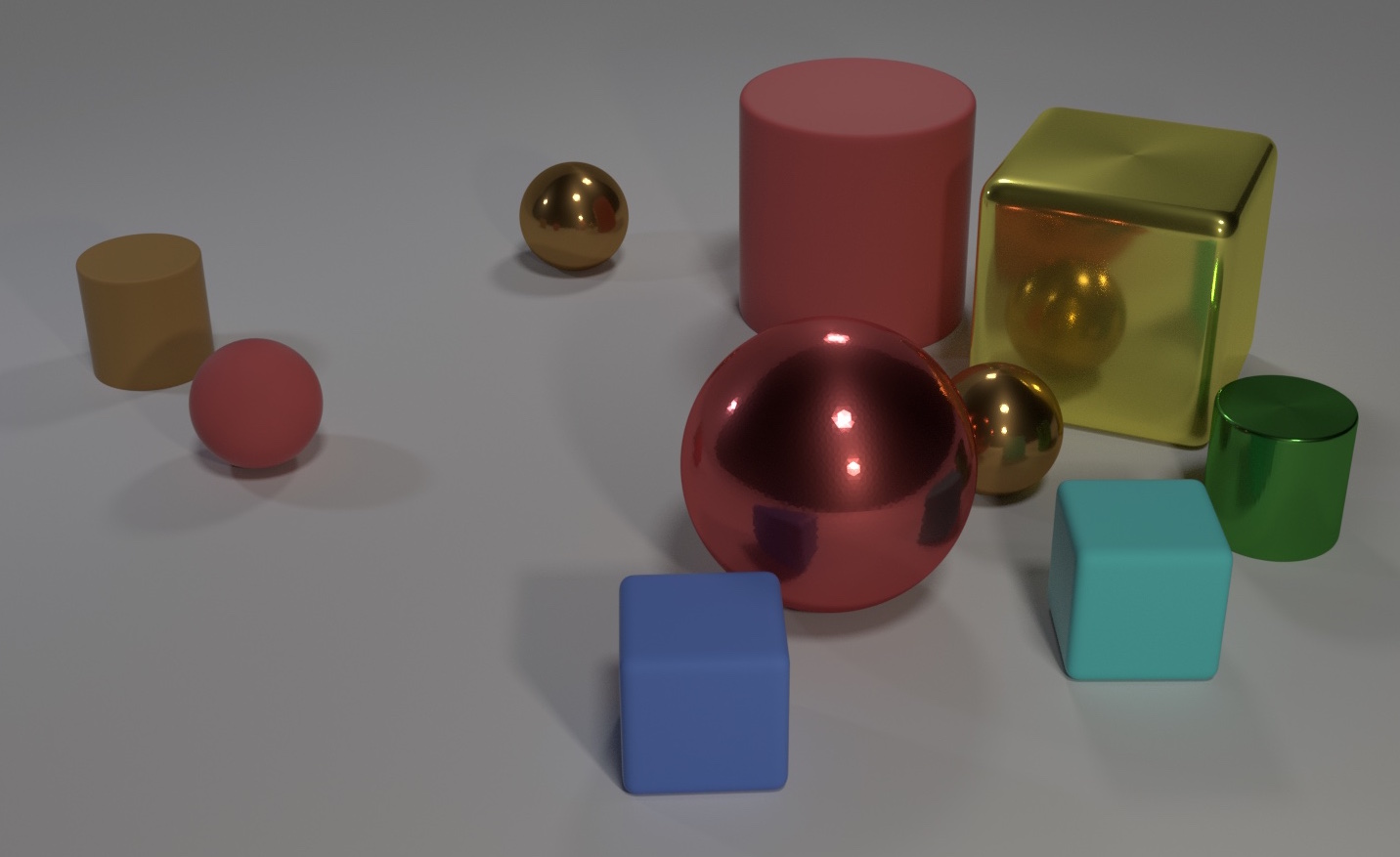}
\begin{lstlisting}[frame=single]
Question: Is there a yellow cube?
\end{lstlisting}
\begin{lstlisting}[frame=single]
Programmatic Query:
  Exists(
    FilterColor(
      FilterShape(Scene(), "cube"),
      "yellow"))
\end{lstlisting}
\begin{lstlisting}[frame=single]
Answer: true
\end{lstlisting}
\caption{A CLEVR example data-point.}
\label{fig:clevr-example}
\end{figure}

We illustrate a concrete example in Fig. \ref{fig:clevr-example}.

\section{Experimental Details}
\label{app:experiment_details}

Our experiments are conducted on a machine with two 20-core Intel Xeon CPUs, four GeForce RTX 2080 Ti GPUs, and 768 GB RAM.
Note that our experiments do not involve training and therefore do not require high-end computation resources.
In this section we elaborate on the foundation models that we used in our experiments and the setup for individual tasks.


\subsection{Model setup}

\paragraph{GPT.}
The default GPT model we use is \inlinecode{gpt-4}.
Depending on the task, there are a few variations we have used which include \inlinescl{gpt-3-turbo}, \inlinecode{text-embedding-ada-002}.
We set the model temperature to $0$ as the default value, and we cache the intermediate result locally for expense-saving purposes.
For chain-of-thought (CoT) prompting, we adopt the zero-shot technique introduced by \cite{kojima2022large}.
Questions that encounter an API server error are manually re-queried.
All experiments are performed from June to August 2023.

\paragraph{OWL-ViT.}
We use the OWLViTProcessor and OwlViTForObjectDetection models from Hugging Face.
We load the pretrained checkpoint \inlinecode{google/owlvit-base-patch32}.
We set the processor's \inlinecode{score\_threshold} to $0.1$, \inlinecode{score\_multiplier} to $1.0$, and 
\inlinecode{expand\_crop\_region} to $0.0$.

\paragraph{ViLT.}
We use the ViltProcessor and ViltForQuestionAnswering models from Hugging Face.
We load the pretrained checkpoint \inlinescl{dandelin/vilt-b32-finetuned-vqa}.
We set the default value of \inlinecode{top} to $5$ and \inlinecode{score\_threshold} to $0.1$.

\paragraph{CLIP.}
We use OpenAI's official implementation with the model set to \inlinescl{ViT-B/32}.

\paragraph{DSFD.}
We use the implementation of DSFD from PyPI's face-detection package.
We set \inlinescl{confidence\_threshold} to 0.5 and \inlinescl{nms\_iou\_threshold} to 0.3.

\paragraph{Segment Anything.}
We use the Segment Anything Model (SAM) from its official open-source repository.
Specifically, we use the ViT-H SAM model checkpoint.
We set the default \inlinescl{iou\_threshold} to 0.88, \inlinescl{area\_threshold} to 0, and \inlinescl{expand\_crop\_region} to 0.

\paragraph{Prompt2prompt and Diffusion Model.}
We adapt Prompt2prompt \cite{hertz2022prompttoprompt} from its official repository to support continuous editing, and we choose the underlying stable diffusion model \inlinescl{CompVis/stable-diffusion-v1-4} from Hugging Face.
We set the default value of \inlinescl{num\_inference\_steps} to 50 for the stable diffusion model, \inlinescl{max\_num\_words} to 77, and \inlinescl{guidance\_scale} to 7.5.


\paragraph{Others.}
The other integrated models which are not used in our experiments includes chat models like Vicuna \cite{zheng2023vicuna}, Llama 2 \cite{touvron2023llama}, and text embedding models like Cross-Encoder \cite{nogueira2019passage}, and RoBERTa \cite{liu2019roberta}.

\subsection{Task setup}

\paragraph{Date reasoning.}
The DR dataset is adapted from BIG-bench's \inlinecode{date understanding} task, with 28 of the original 369 questions being corrected for wrong target answers.
We solve this task by decomposing it into two sub-tasks: extracting structured information using an LM, followed by logical reasoning over the structured information using relational rules and date arithmetic.

For the first sub-task, we leverage GPT-4 with 5-shot prompting for extracting the following three relations from the given context:
\begin{enumerate}
    \item
    \inlinescl{mentioned\_date(label, date)}: \inlinescl{label} is a string label for a date whose MM/DD/YYYY form is explicitly mentioned in the context, and \inlinescl{date} is the corresponding MM/DD/YYYY string.
    \item
    \inlinescl{goal(label)}: \inlinescl{label} is the date label whose MM/DD/YYYY form is requested as the answer.
    \item
    \inlinescl{relationship(earlier\_date, later\_date, diff)}: the first two arguments are a pair of date labels relevant to the question, and \inlinescl{diff} is the time duration between the dates.
\end{enumerate}

See Table \ref{tab:dr-ex} for an example set of extracted relations.
The shots for \inlinescl{gpt\_extract\_relation} are manually composed to be similar to questions in the dataset.
For this task specifically, we configure \inlinescl{gpt\_extract\_relation} to use zero-shot CoT \cite{kojima2022large} when extracting \inlinescl{relationship}, which improves accuracy by over 10\%.

After extracting these relations, a symbolic program iterates through derived dates and durations to compute dates for all extracted date labels, including the goal date.
Date parsing and date arithmetic is enabled by \tool{}'s built-in data types \inlinescl{DateTime} and \inlinescl{Duration}.

\begin{table}
  \centering
  \scriptsize
  \begin{tabular}{
    >{\raggedleft\arraybackslash}m{0.11\textwidth}
    >{\arraybackslash}m{0.31\textwidth}} 
    \toprule
     \multicolumn{2}{m{0.42\textwidth}}{\textbf{Question:} Today's meeting is rescheduled to 11 am tomorrow, 10/16/1924. What is the date one year ago from today in MM/DD/YYYY?} \\ 
    \midrule
    \textbf{\tool{} extracted relations:} & \tiny
      \textbf{mentioned\_date:} [("rescheduled-meeting", "1924-10-16")] \newline
      \textbf{relationship:} [("1-year-ago", "today", "R12MO PT0S"), ("today", "rescheduled-meeting", "P1D")] \newline
      \textbf{goal:} [("1-year-ago")] \\
    \midrule
    \textbf{\tool{} answer:} & 10/15/1923 \textcolor{Green}{(CORRECT)} \\
    \midrule
    \textbf{GPT-4 response:} & \tiny Let's think step by step. First, we need to subtract one year from the given date, 10/16/1924. When we do this, we get 10/16/1923. \\
    \midrule
    \textbf{GPT-4 answer:} & 10/16/1923 \textcolor{Red}{(INCORRECT)} \\
    \bottomrule
  \end{tabular}
  \caption{
    Comparison between our solution and GPT-4 (zero-shot CoT) on selected question from DR dataset.
  }
  \label{tab:dr-ex}
\end{table}

\paragraph{Tracking shuffled objects.}
The TSO dataset is randomly sampled from a combined dataset of subtasks corresponding to $n=3,5,7$ objects from BIG-bench's \inlinecode{tracking shuffled objects} task.
Specifically, our random sample contains 32 questions where $n=3$, 59 questions where $n=5$, and 59 questions where $n=7$.
Our solution relies on GPT-4 with single-shot prompting for extracting three relations:
\begin{enumerate}
    \item
    \inlinescl{possessions(time, person, object)}: \inlinescl{person} possesses \inlinescl{object} at time step \inlinescl{time}.
    We prompt GPT-4 to only extract the initial possessions (where \inlinescl{time} is 1), which are explicitly described in the context.
    \item
    \inlinescl{swaps(time, person\_a, person\_b)}: \inlinescl{person\_a} and \inlinescl{person\_b} swap objects at time step \inlinescl{time}.
    \item
    \inlinescl{goal(person)}: \inlinescl{person} is the target person whose final possessed object is expected as the answer.
\end{enumerate}

See Table \ref{tab:tso-ex} for an example set of extracted relations.
We prompt \inlinescl{gpt\_extract\_relation} with one shot based on a question from the BIG-bench task but not from our sampled dataset.
Our reasoning program then iterates through all the swaps starting from the initial possessions and retrieves the last possessed object associated with the target.

We conjecture that the exemplary performance of our model on TSO is due to the highly consistent syntactic structure of the NL inputs, facilitating relation extraction under a one-shot setting.

\begin{table}
  \centering
  \scriptsize
  \begin{tabular}{
    >{\raggedleft\arraybackslash}m{0.11\textwidth}
    >{\arraybackslash}m{0.31\textwidth}} 
    \toprule
    \multicolumn{2}{m{0.42\textwidth}}{\textbf{Question:} Alice, Bob, Claire, Dave, and Eve are dancers at a square dance. At the start of a song, they each have a partner: Alice is dancing with Ophelia, Bob is dancing with Lola, Claire is dancing with Izzi, Dave is dancing with Karl, and Eve is dancing with Jamie. Throughout the song, the dancers often trade partners. First, Bob and Claire switch partners. Then, Eve and Dave switch partners. Then, Eve and Claire switch partners. Then, Alice and Eve switch partners. Finally, Alice and Bob switch partners. At the end of the dance, Bob is dancing with} \\ 
    \midrule
    \textbf{\tool{} extracted relations:} & \tiny
      \textbf{possessions:} {\tiny [(1, "Alice", "Ophelia"), (1, "Bob", "Lola"), (1, "Claire", "Izzi"), (1, "Dave", "Karl"), (1, "Eve", "Jamie")]} \newline
      \textbf{swaps:} {\tiny [(1, "Bob", "Claire"), (2, "Eve", "Dave"), (3, "Eve", "Claire"), (4, "Alice", "Eve"), (5, "Alice", "Bob")]} \newline
      \textbf{goal:} {\tiny [("Bob")]} \\
    \midrule
    \textbf{\tool{} answer:} & Lola \textcolor{Green}{(CORRECT)} \\
    \midrule
    \textbf{GPT-4 response:} & \tiny Let's think step by step. At the start, Bob is dancing with Lola. After Bob and Claire switch partners, Bob is dancing with Izzi. After Eve and Dave switch partners, Bob is still dancing with Izzi. After Eve and Claire switch partners, Bob is still dancing with Izzi. After Alice and Eve switch partners, Bob is still dancing with Izzi. Finally, after Alice and Bob switch partners, Bob is dancing with Ophelia. So, the answer is (A) Ophelia. \\
    \midrule
    \textbf{GPT-4 answer:} & Ophelia \textcolor{Red}{(INCORRECT)} \\
    \bottomrule
  \end{tabular}
  \caption{
    Comparison between our solution and GPT-4 (zero-shot CoT) on selected question from TSO dataset.
  }
  \label{tab:tso-ex}
\end{table}

\paragraph{Kinship reasoning.}
We use the CLUTRR \cite{sinha2019clutrr} dataset to perform the kinship reasoning task in NLP. 
Each data point in this task contains a story that indicates the kinship between characters, and the aim is to infer the relationship between two specified characters. 
We use all the 1,146 data points in the test dataset of 089907f8.

The intermediate symbolic representation we use GPT-4 model to extract is a kinship graph, whose edge is composed of 20 different relationships, including ``father'', ``mother'', ``uncle'', ``niece''.
For prompting the GPT-4 model, we first ask the GPT model to yield us all the kinships relations that are mentioned in the context and store it in \inlinecode{mentioned\_kinship}.
Then we also need to extract the two target characters, in which we consult gpt and store the answer in the relation \inlinecode{query}.
See Table \ref{tab:kr-ex} for an example set of extracted relations.

The resulting kinship graph is then reasoned along with a given external knowledge base, which includes the compositional knowledge like ``father's mother is grandmother'', to obtain the relationship between the two desired people.

\begin{table}
  \centering
  \scriptsize
  \begin{tabular}{
    >{\raggedleft\arraybackslash}m{0.11\textwidth}
    >{\arraybackslash}m{0.31\textwidth}} 
    \toprule
    \multicolumn{2}{m{0.42\textwidth}}{\textbf{Question:} Myrna and her husband Christopher went on a cruise. They had a wonderful time. Christopher and his daughter Lucille took a day off school to go to the zoo. Who is Lucille to Myrna?} \\
    \midrule
    \textbf{\tool{} extracted relations:} & \tiny
      \textbf{mentioned\_kinship:} [("Christopher",
          "Lucille",
          "father"),
        ("Christopher",
          "Myrna",
          "husband"),
        ("Lucille",
          "Christopher",
          "daughter"),
        ("Myrna",
          "Christopher",
          "wife")] \newline
      \textbf{query:} [("Lucille", "Myrna")] \\
    \midrule
    \textbf{\tool{} answer:} & daughter \textcolor{Green}{(CORRECT)} \\
    \midrule
    \textbf{GPT-4 response:} & \tiny Lucille is Myrna's stepdaughter. In the given scenario, Christopher is Myrna's husband, and Christopher is also the father of Lucille. This makes Lucille the daughter of Myrna's husband, which means she is Myrna's stepdaughter. \\
    \midrule
    \textbf{GPT-4 answer:} & stepdaughter \textcolor{Red}{(INCORRECT)} \\
    \bottomrule
  \end{tabular}
  \caption{
    Comparison between our solution and GPT-4 (zero-shot CoT) on selected question from CLUTRR dataset.
  }
  \label{tab:kr-ex}
\end{table}

\paragraph{Math reasoning.}

This task is drawn from the GSM8K dataset of arithmetic word problems \cite{cobbe2021training}.
Both our math and date reasoning datasets have previously served as benchmarks for LLM performance under chain-of-thought prompting \cite{wei2023chainofthought, kojima2022large}.
The questions involve grade school math word problems created by human problem writers, and the model is asked to produce a number as the result.
Since the output can be fractional, we allow a small delta when comparing the derived result with the ground truth.
We include GPT-4 (few-shot and CoT) as baselines for this task.

Our solution to this task prompts GPT-4 (2-shot) to produce step-by-step expressions, which can contain constants, variables, and simple arithmetic operations.
For example, the fact \inlinescl{assign("total\_sale", "april\_sale + may\_sale")} represents that total sales are the sum of April and May sales.
See Table \ref{tab:mr-ex} for an example set of extracted steps.

We evaluate all the expressions through a DSL, and the result associated with the goal variable is returned.
By focusing the LM's responsibility solely on semantic parsing, our relational program can then achieve faithful numerical computation via DSL evaluation.

The semantic parsed result is a Python expression that we can directly call the \inlinecode{eval} function in Python over the string, and we can obtain the desired outcome. 
This \inlinecode{py\_eval} function is wrapped as a foreign attribute in \todo{}.

\begin{table}
  \centering
  \scriptsize
  \begin{tabular}{
    >{\raggedleft\arraybackslash}m{0.11\textwidth}
    >{\arraybackslash}m{0.31\textwidth}} 
    \toprule
    \multicolumn{2}{m{0.42\textwidth}}{\textbf{Question:} If it takes 3 kangaroos traveling at the same speed a total of 18 hours to travel across a highway, how many hours will it take four turtles, each traveling at half the speed of a kangaroo, to do so?} \\ 
    \midrule
    \textbf{\tool{} extracted steps:} & \tiny
       (0, "kangaroo\_speed", "1 / (18 / 3)"), \newline
       (1, "turtle\_speed", "\{kangaroo\_speed\} / 2"), \newline
       (2, "turtle\_time", "1 / \{turtle\_speed\}"), \newline
       (3, "total\_turtle\_time", "\{turtle\_time\} * 4") \\
    \midrule
    \textbf{\tool{} answer:} & 48 \textcolor{Green}{(CORRECT)} \\
    \midrule
    \textbf{GPT-4 response:} & \tiny Let's think step by step. First, we know that the total time it takes for 3 kangaroos to travel across the highway is 18 hours. This means that one kangaroo takes 18 hours to travel across the highway. Then, we know that a turtle travels at half the speed of a kangaroo, which means it takes twice as long for a turtle to travel the same distance. So, one turtle would take 18 hours * 2 = 36 hours to travel across the highway. However, we have four turtles traveling together at the same speed, so they would still take 36 hours to travel across the highway, because they are traveling simultaneously. \\
    \midrule
    \textbf{GPT-4 answer:} & 36 \textcolor{Red}{(INCORRECT)} \\
    \bottomrule
  \end{tabular}
  \caption{
    Comparison between our solution and GPT-4 (zero-shot CoT) on selected question from GSM8k dataset.
  }
  \label{tab:mr-ex}
\end{table}

\paragraph{Retrieval augmentation and semantic search.}

We have two benchmarks for retrieval augmentation and semantic search: HotpotQA and Amazon's ESCI Product Search.

The HotpotQA \cite{yang2018hotpotqa} data, a Wikipedia-based question answering (QA) dataset under the ``distractor" setting.
Here, the model takes in $2$ parts of inputs:
1) a question, and
2) $10$ Wikipedia paragraphs as the context for answering the question.
Among the $10$ Wikipedia pages, at most $2$ are relevant to the answer, while the others are distractors.
This challenges the capability of retrieving information based on the question.
Since the QA models produce free-form answers that can vary largely, we use GPT-4 to check the correctness of the derived result against the ground truth.
This is aided by the manual inspection of subsets to determine the statistical variance.

We use Amazon's ESCI Product Search dataset \cite{reddy2022shopping}.
The model is provided with a natural language (NL) query and a list of products ($23$ products on average).
The goal is to rank the products that best match the query.
In the dataset, for each pair of query and product, a label among $E$ (exact match), $S$ (substitute), $C$ (complementary), and $I$ (irrelevant) is provided.
The metric we use to evaluate the performance is nDCG.
The gains are set to be $1.0$ for $E$, $0.1$ for $S$, $0.01$ for $C$, and $0.0$ for $I$.
We include GPT's embedding model, \inlinescl{text-embedding-ada-002} as baselines for ranking products.

\paragraph{Compositional multi-modal reasoning.}

For compositional multi-modal reasoning, we pick tasks of CLEVR and GQA.
We choose two compositional VQA datasets, GQA \cite{drew2019gqa} and CLEVR \cite{johnson2016clevr}.
In this task, the model is given an image and a question, and needs to answer the question.
For GQA, the majority of questions expect yes/no answers, while CLEVR's questions demand features like counting and spatial reasoning.
We randomly sample 184 and 480
The images and questions in GQA are collected from real life while that of CLEVR are synthetic.

\paragraph{Visual object tagging.}
For VQAR, we consider the top 50 object bounding boxes returned by OWL-ViT.
Our relational knowledge base is from \cite{huang2021scallop}.
When querying ViLT, we take the top response from a score threshold of 0.9.
We manually score semantic correctness by finding the percentage of objects returned that match the query.
Object bounding boxes are considered correct if they contained any part of an entity matching the query.

For OFCP, we curated 50 examples featuring groups of notable celebrities and politicians from Wikimedia Commons and other Internet sources, and manually assigned descriptive filenames to each image.
We obtain the set of possible names by prompting GPT-4 with the filename.
We enlarge the face bounding boxes returned by DSFD by a factor of 1.3 before querying CLIP.
We tag each face with its most probable name from CLIP, but if the probability is below the 0.8 threshold, then the face is tagged ``unknown''.
The ground truth of relevant faces and their names were manually assigned based on the filename description.
The ground truth label for non-relevant faces is ``unknown''.
All faces judged to be in the foreground of an image, as well as any additional faces not tagged with ``unknown'', are counted for semantic correctness.

\paragraph{Image generation and editing.}
We manually wrote $20$ prompts image generation and their editing sequences. 
Each prompt includes one image generation prompt and two consequent image editions. 
Our domain-specific language for image generation and editing supports $5$ operations: \inlinecode{Background}, \inlinecode{ReplaceObject}, \inlinecode{RefineObject}, \inlinecode{NotObject}, \inlinecode{ReweightObject}.
We use the GPT-4 model to convert the natural language prompts into programmatic queries with 4 shot examples. 
There are $2$ cases among $20$ that fail to convert the natural language into executable programs, as the \inlinecode{Replace} operation requires to have the same token length of the original input text and the updated text, while the GPT-4 model fails to capture the requirement through the few-shot examples.

\section{Qualitative Studies}
\label{app:qualitative_studies}
\begin{figure*}
    \centering
    \captionsetup{font=footnotesize}
    \begin{subfigure}{0.35\linewidth}
        \centering
        \includegraphics[height=4cm]{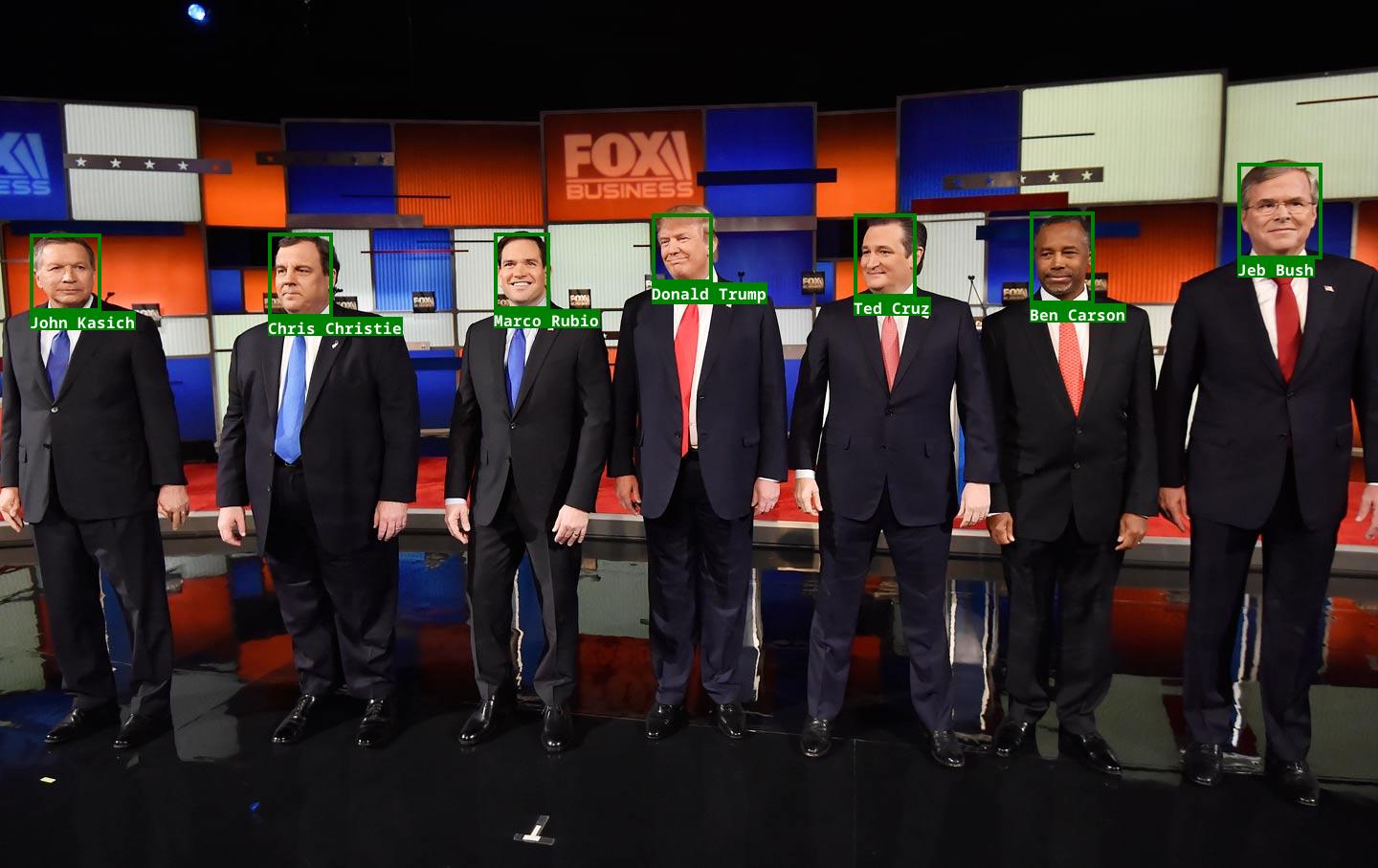}
        \caption{2016\_GOP\_Debate\_SC\_ap\_img.jpg}
    \end{subfigure}
    \hfill
    \begin{subfigure}{0.35\linewidth}
        \centering
        \includegraphics[height=4cm]{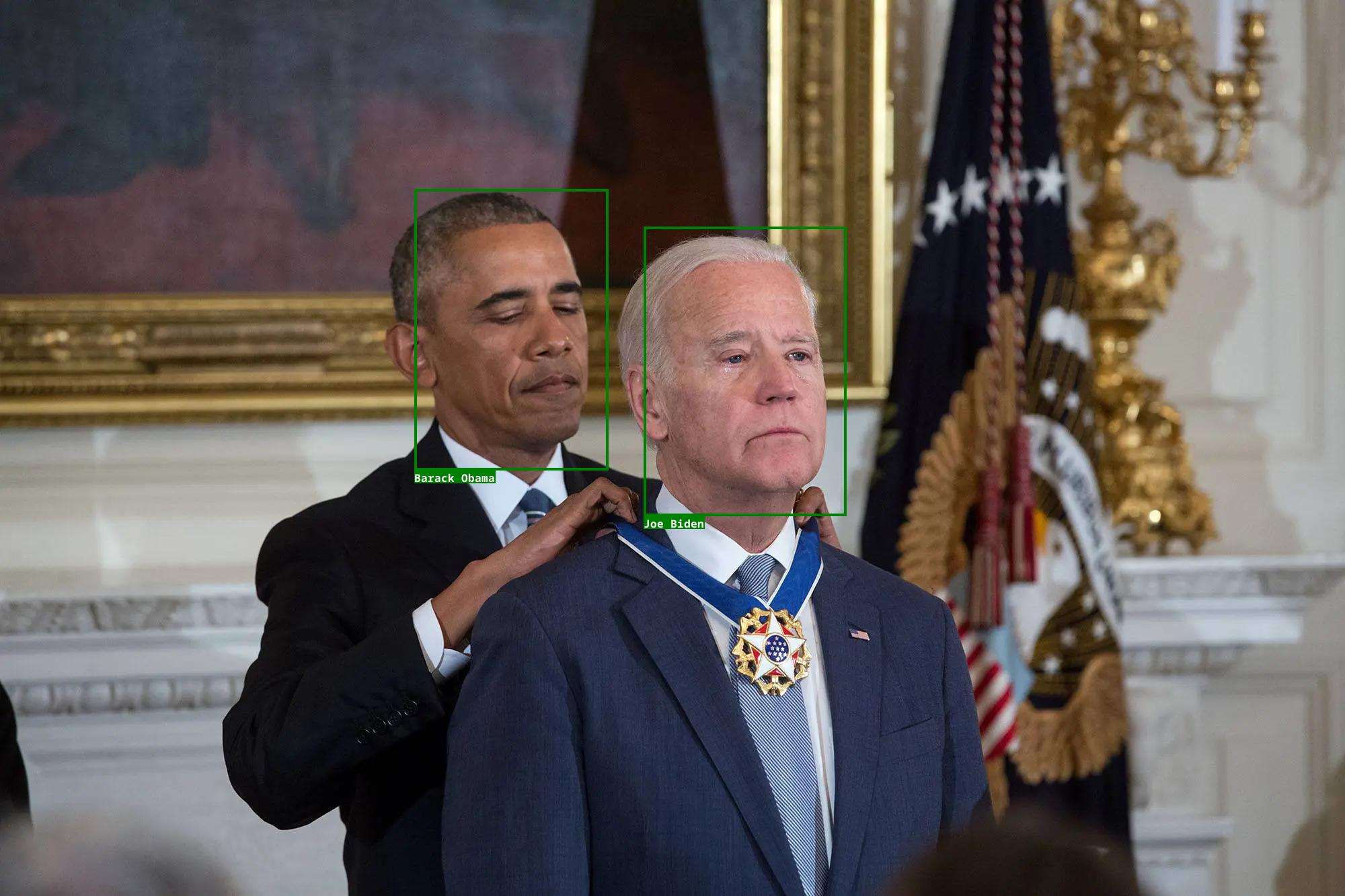}
        \caption{Joe\_Biden\_Receives\_Presidential\_Medal\_of\_Freedom.jpg}
    \end{subfigure}
    \hfill
    \begin{subfigure}{0.25\linewidth}
        \centering
        \includegraphics[height=4cm]{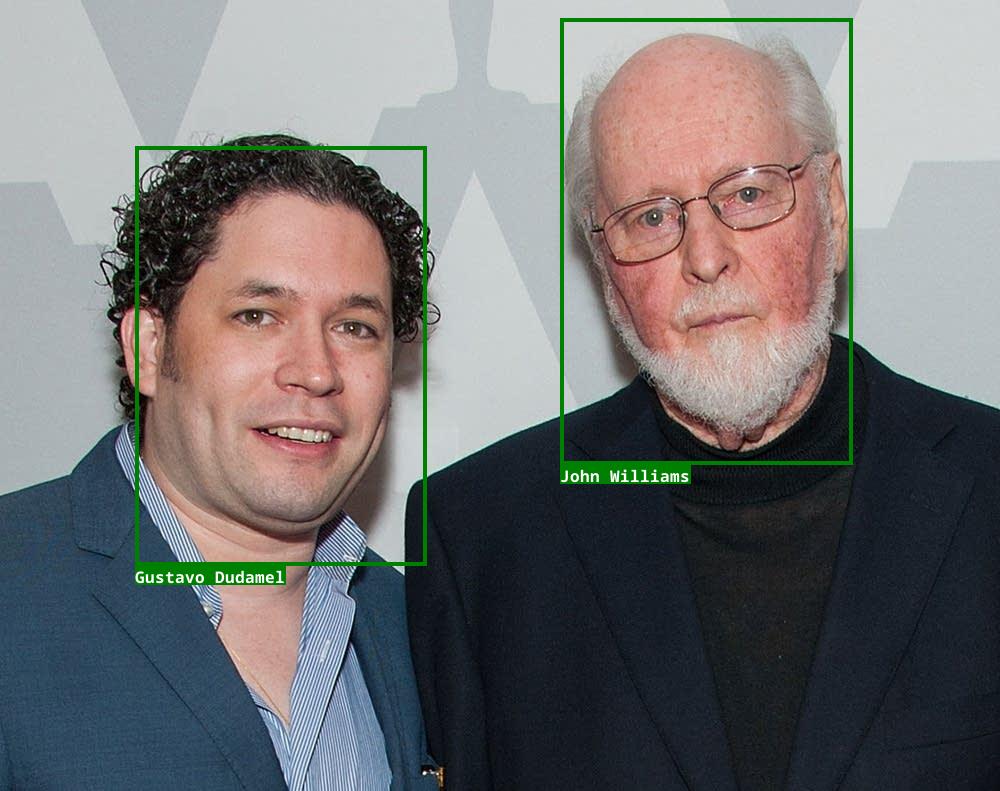}
        \caption{dudamel\_williams.jpg}
    \end{subfigure}

    \vspace{0.5cm}
    \begin{subfigure}{0.35\linewidth}
        \centering
        \includegraphics[height=4cm, width=6.4cm]{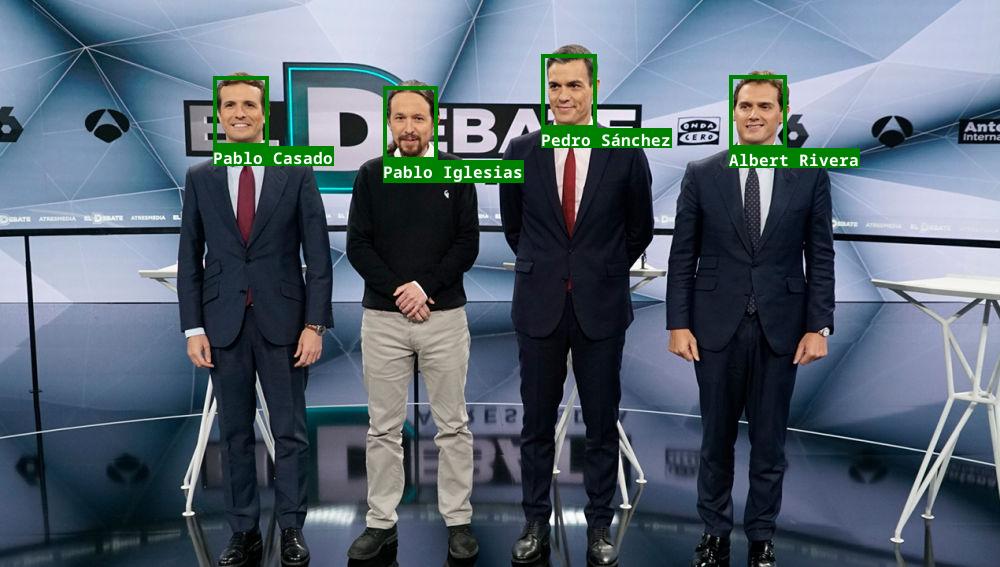}
        \caption{2019 Spanish General Election Debate.jpg}
    \end{subfigure}
    \hfill
    \begin{subfigure}{0.32\linewidth}
            \centering
        \includegraphics[height=4cm]{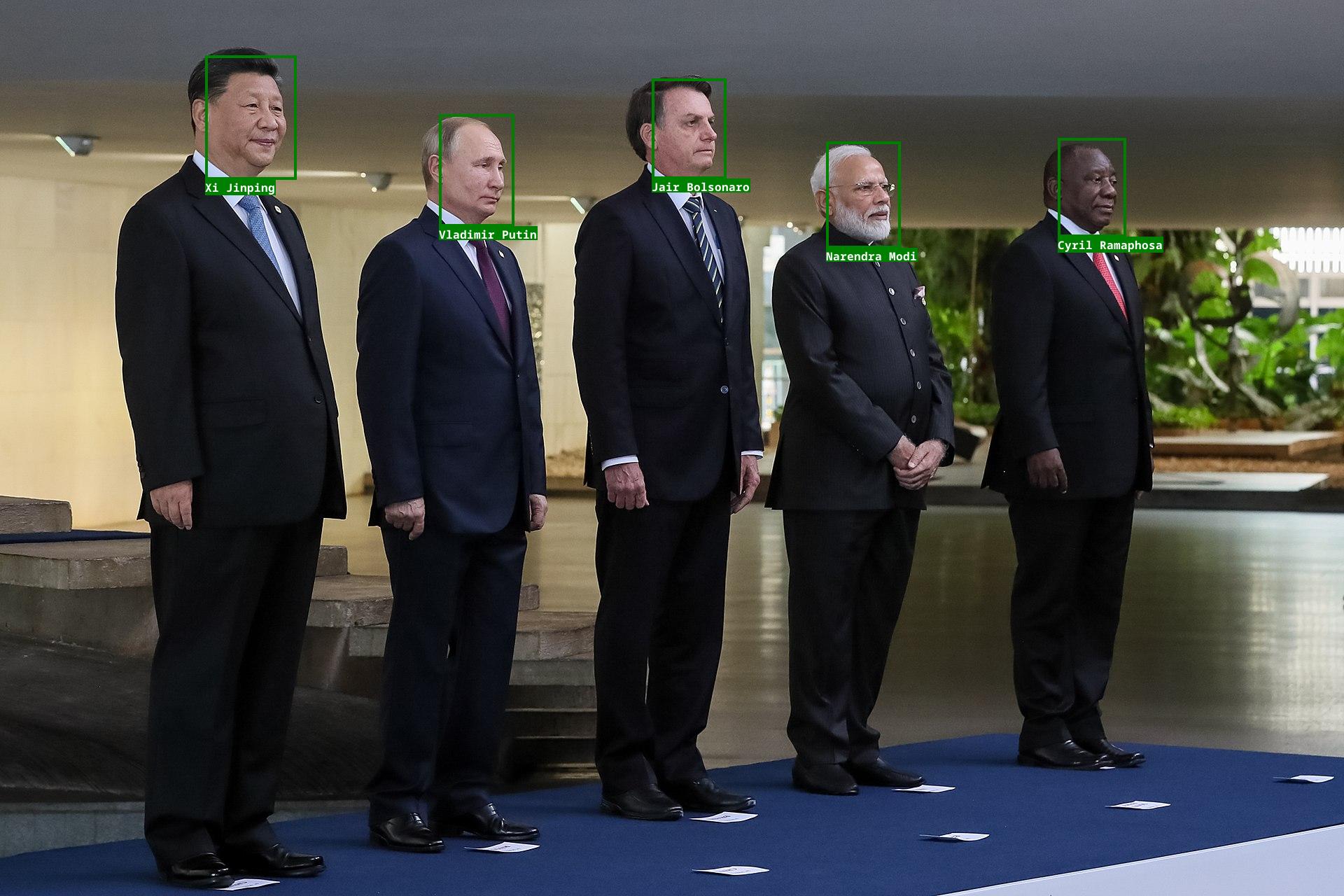}
        \caption{BRICS Summit 2019.jpg}  
    \end{subfigure}
    \hfill
    \begin{subfigure}{0.25\linewidth}
        \centering
        \includegraphics[height=4cm, width=5cm]{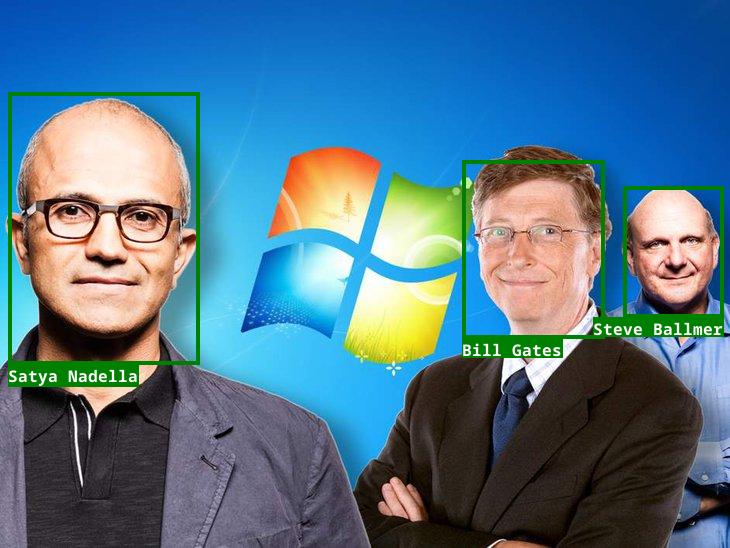}
        \caption{microsoft ceos.jpeg}
    \end{subfigure}
    \captionsetup{font=normalsize}
    \caption{Face Tagging (OFCP) exemplars.}
    \label{fig:face_tagging}
\end{figure*}

\begin{figure*}
    \centering
    \captionsetup{font=footnotesize}
    \begin{subfigure}{0.33\linewidth}
        \centering
        \includegraphics[height=4cm, width=5.7cm]{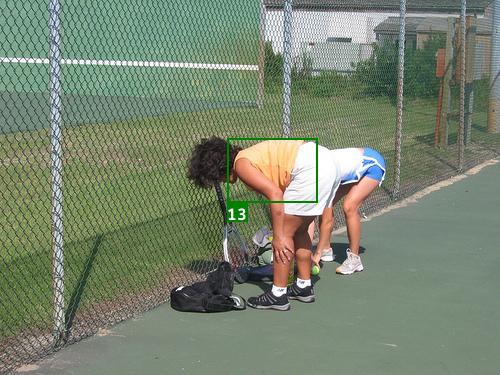}
\begin{lstlisting}[frame=single]
Q: Find the orange objects.
\end{lstlisting}
\begin{lstlisting}[frame=single]
Programmatic Query: 
    Find_Attr("orange")
\end{lstlisting}
\begin{lstlisting}[frame=single]
Answer: {13}
\end{lstlisting}

    \end{subfigure}
    \hfill
    \begin{subfigure}{0.66\linewidth}
        \centering
        \includegraphics[height=4cm, width=5.7cm]{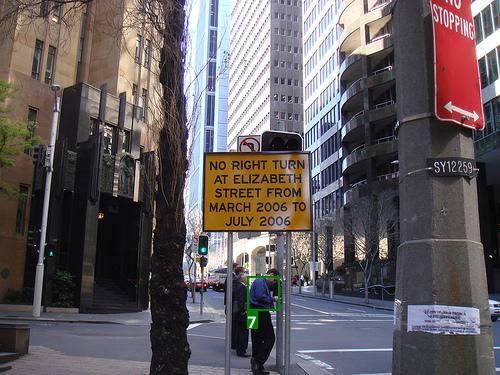}
        \hfill
        \includegraphics[height=4cm, width=5.7cm]{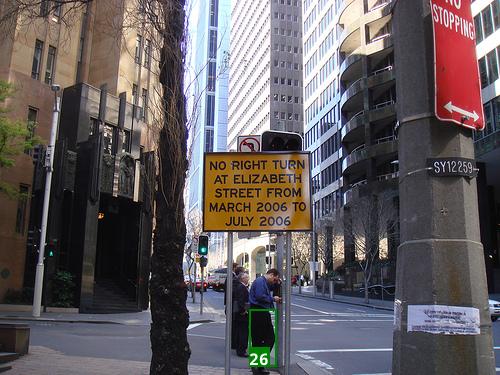}
\begin{lstlisting}[frame=single]
Question: Find the objects that are clothing. 
\end{lstlisting}
\begin{lstlisting}[frame=single]
Programmatic Query: 
    Hypernym_Find("clothing")
\end{lstlisting}
\begin{lstlisting}[frame=single]
Answer: {7, 26}
\end{lstlisting}
\end{subfigure}

    \vspace{0.5cm}
    \begin{subfigure}{0.33\linewidth}
        \centering
        \includegraphics[height=4cm, width=5.7cm]{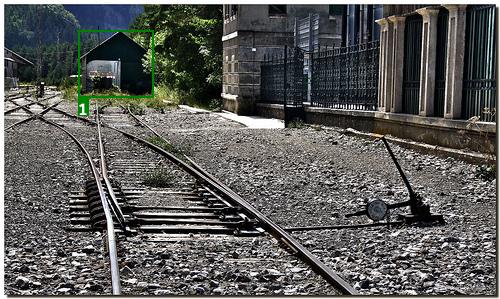}
\begin{lstlisting}[frame=single]
Q: Objs used for housing family.
\end{lstlisting}
\begin{lstlisting}[frame=single]
Programmatic Query: 
    KG_Find(X, "used for", 
            "housing family")
\end{lstlisting}
\begin{lstlisting}[frame=single]
Answer: {1}
\end{lstlisting}
    \end{subfigure}
    \hfill
    \begin{subfigure}{0.66\linewidth}
        \centering
        \includegraphics[height=4cm, width=5.7cm]{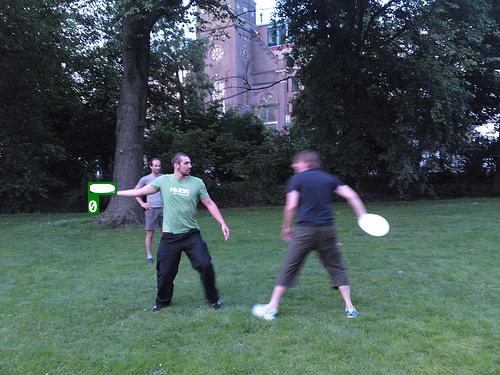}
        \hfill
        \includegraphics[height=4cm, width=5.7cm]{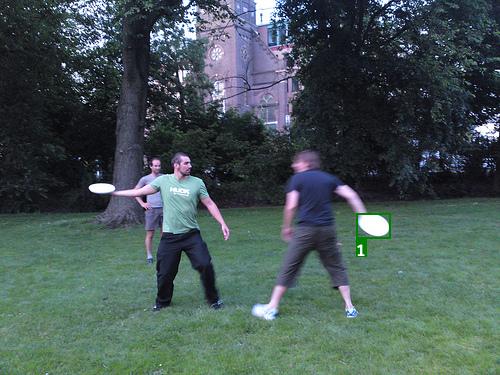}
\begin{lstlisting}[frame=single]
Question: Find the objects that are sports equipment.
\end{lstlisting}
\begin{lstlisting}[frame=single]
Programmatic Query: 
    Hypernym_Find(
       "sports equipment")
\end{lstlisting}
\begin{lstlisting}[frame=single]
Answer: {0, 1}
\end{lstlisting}
       
    \end{subfigure}
    \caption{Object Tagging (VQAR) exemplars.}
    \label{fig:object_tagging}
\end{figure*}

\begin{figure*}
    \centering
    \captionsetup{font=footnotesize}
    \begin{subfigure}{0.33\linewidth}
        \centering
        \includegraphics[trim={0 25cm 0 5cm}, clip, height=4cm, width=5.7cm]{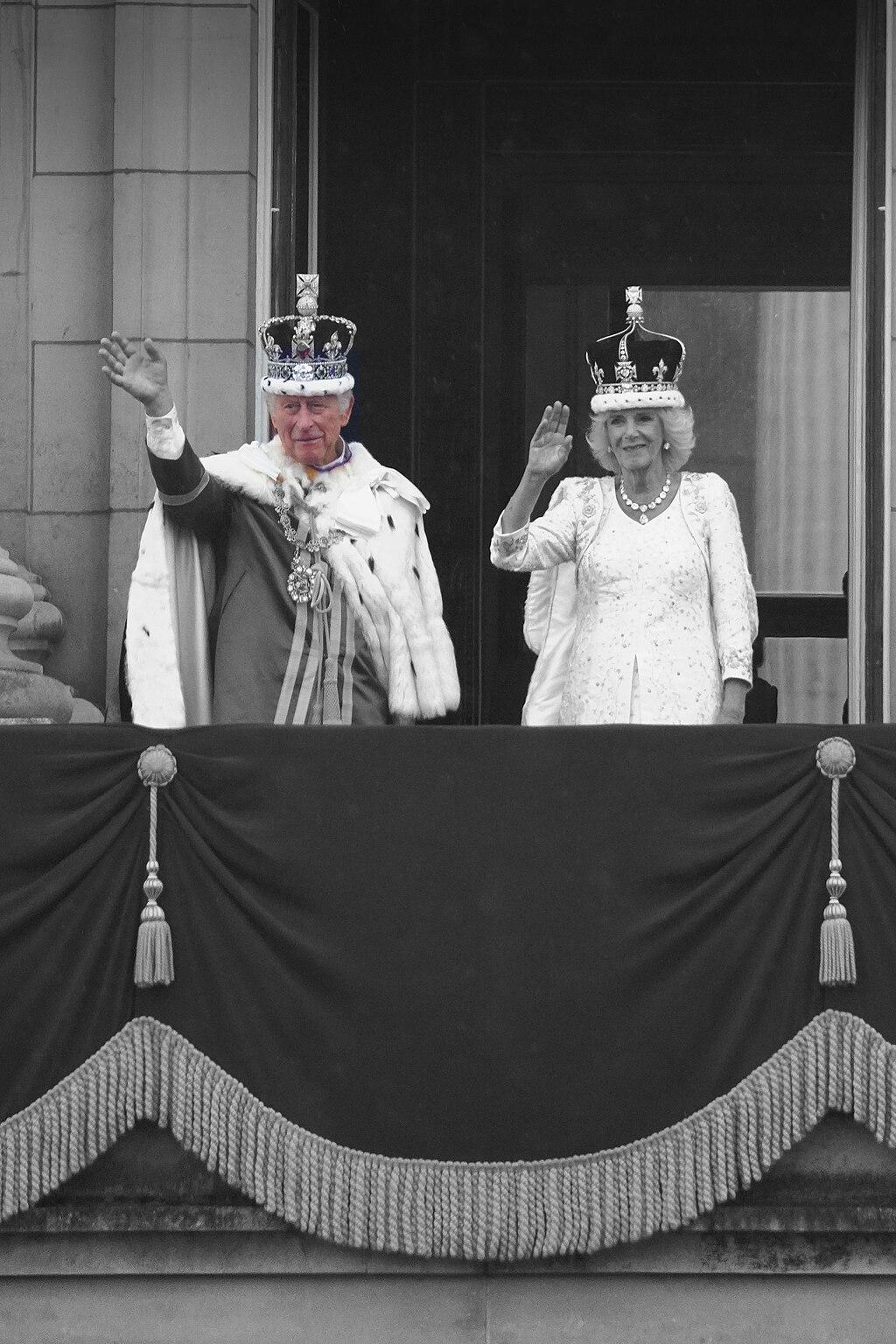}
\begin{lstlisting}[frame=single]
Instruction: 
Do a color pop of 
the man in purple.
\end{lstlisting}
\end{subfigure}
\hfill
\begin{subfigure}{0.33\linewidth}
\centering
\includegraphics[height=4cm, width=5.7cm]{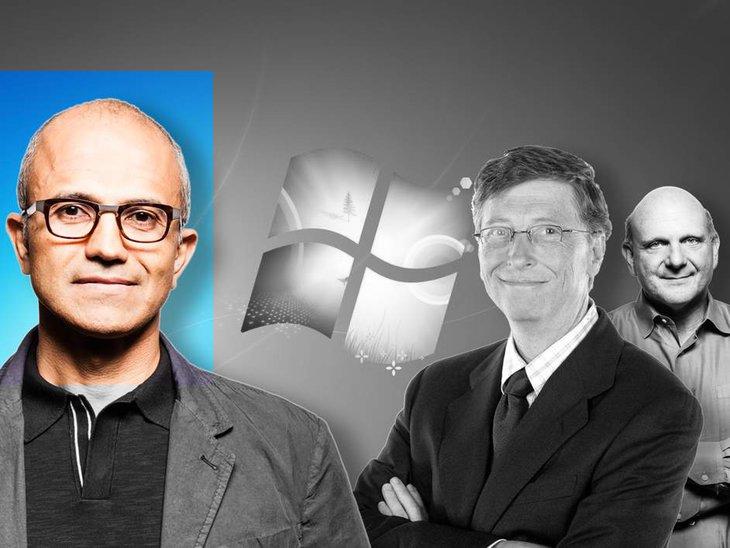}
\begin{lstlisting}[frame=single]
Instruction: 
Make a color pop of 
the current Microsoft CEO.
\end{lstlisting}
\end{subfigure}
\begin{subfigure}{0.33\linewidth}
        \centering
        \includegraphics[height=4cm, width=5.7cm]{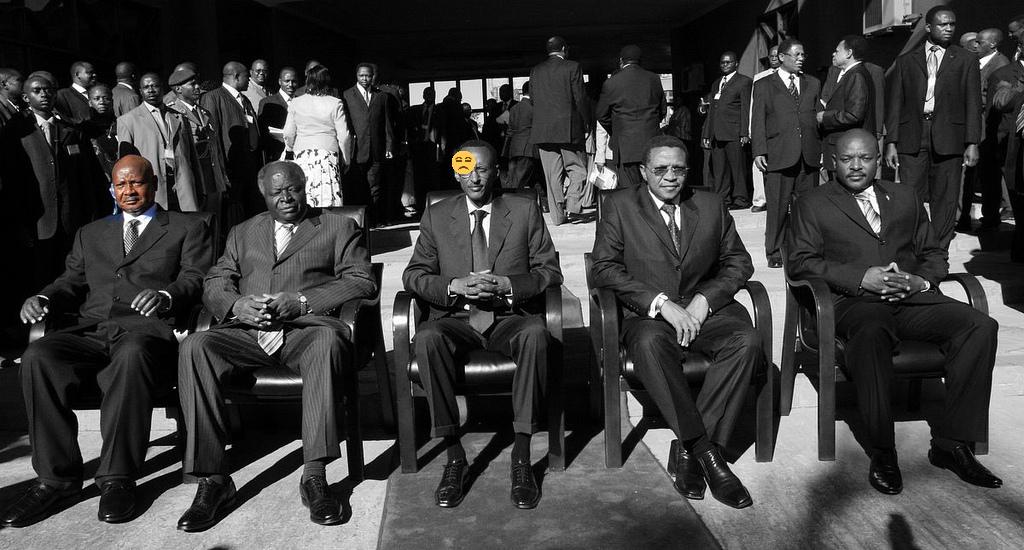}
\begin{lstlisting}[frame=single]
Instruction: Create a color pop 
of the Ugandan President and put 
unamused_face over Kagame.
\end{lstlisting}
\end{subfigure}
\hfill

\begin{subfigure}{0.33\linewidth}
\centering
        \includegraphics[trim={0cm 10cm 0cm 0cm}, clip, height=4cm, width=5.7cm]{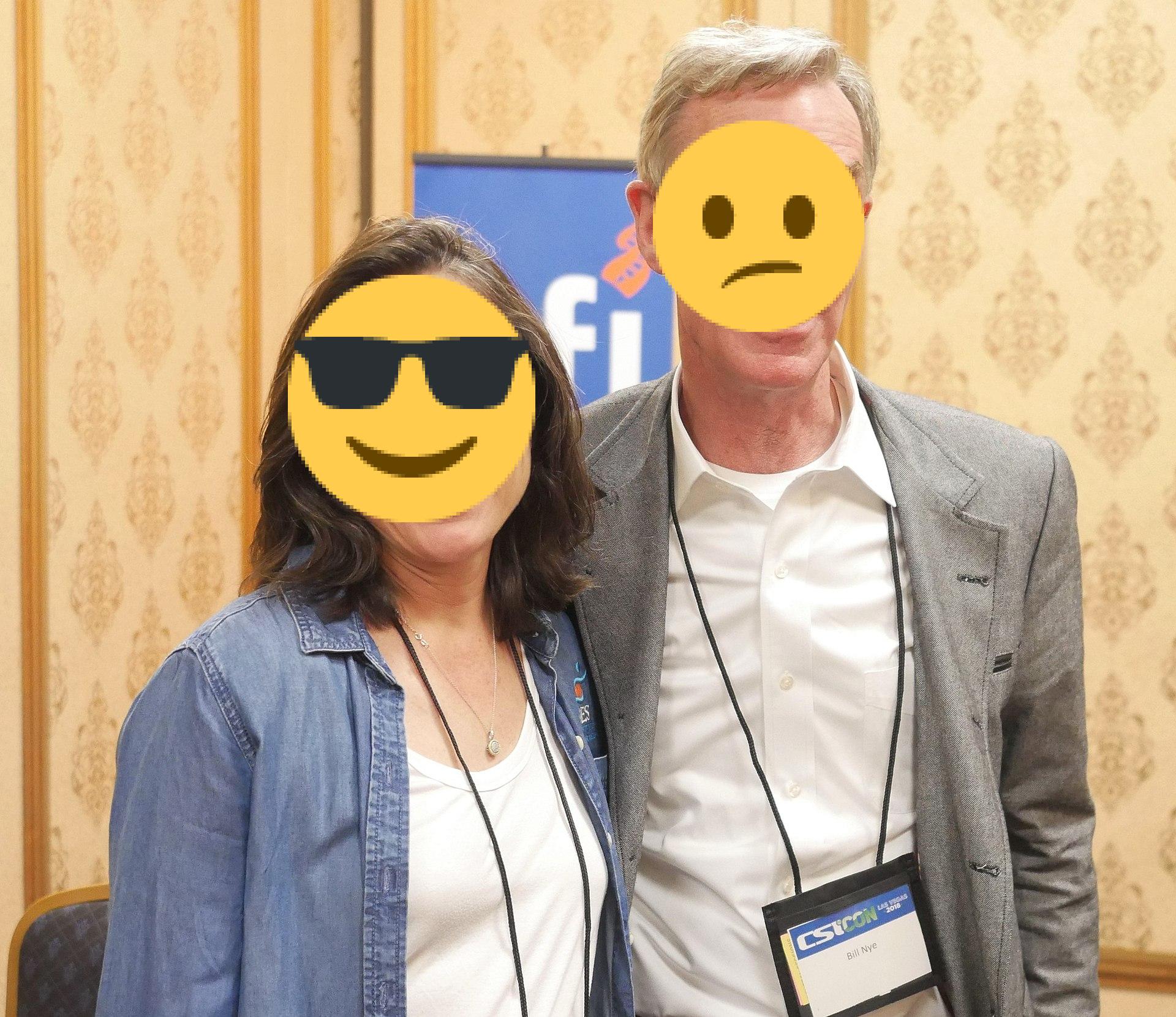}
        \hfill
\begin{lstlisting}[frame=single]
Instruction: 
Cover Bertha Vasquez with 
smiling_face_with_sunglasses 
and Bill Nye with confused_face.
\end{lstlisting}
\end{subfigure}
\hfill
\begin{subfigure}{0.33\linewidth}
\centering
\includegraphics[trim={0cm 15cm 0cm 5cm}, clip, height=4cm, width=5.7cm]{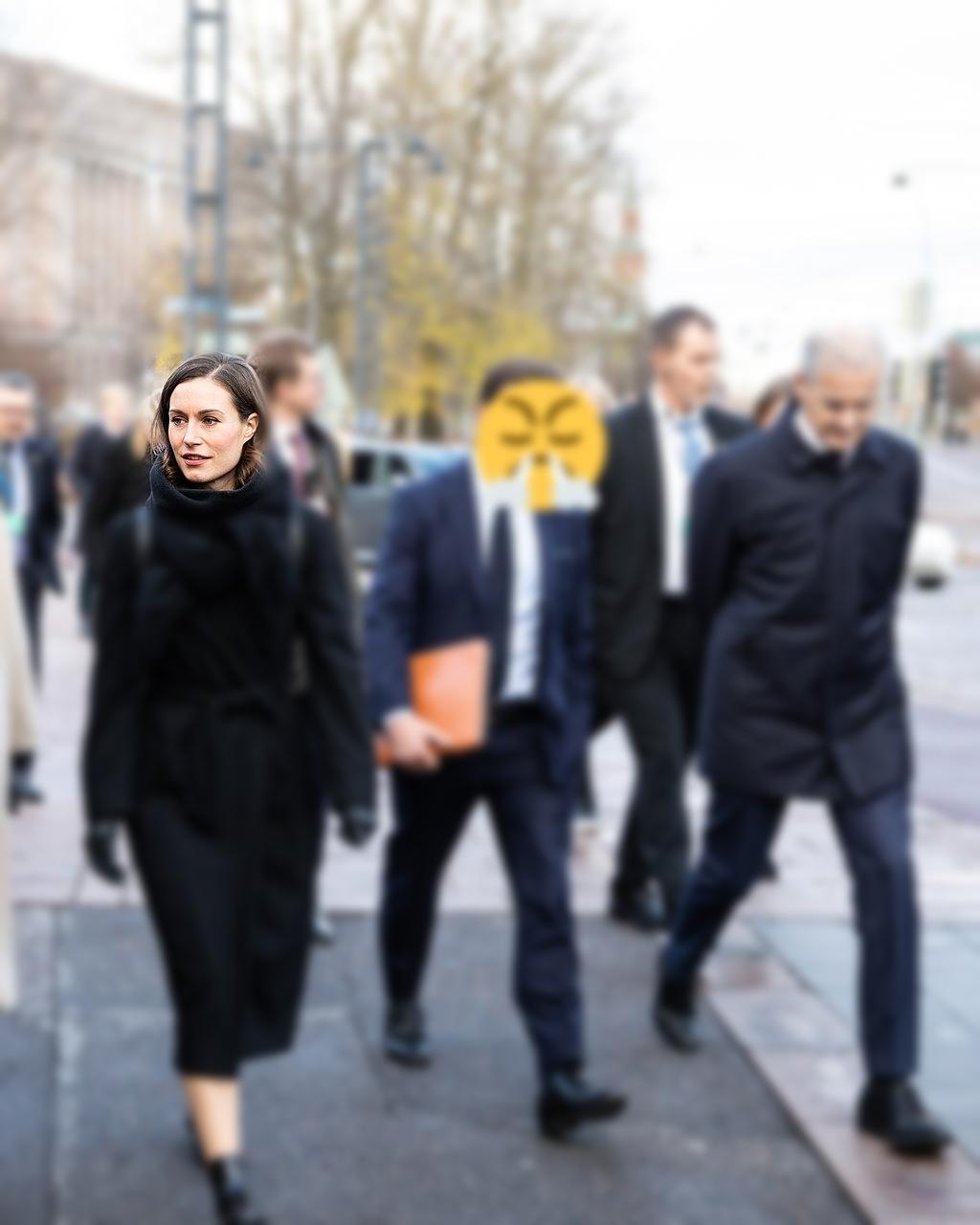}
\hfill
\begin{lstlisting}[frame=single]
Instruction: Hide Ulf Kristersson 
with face_with_steam_from_nose 
and blur everyone except Sanna 
Marin.
\end{lstlisting}
\end{subfigure}
\hfill
\begin{subfigure}{0.33\linewidth}
\centering
\includegraphics[height=4cm, width=5.7cm]{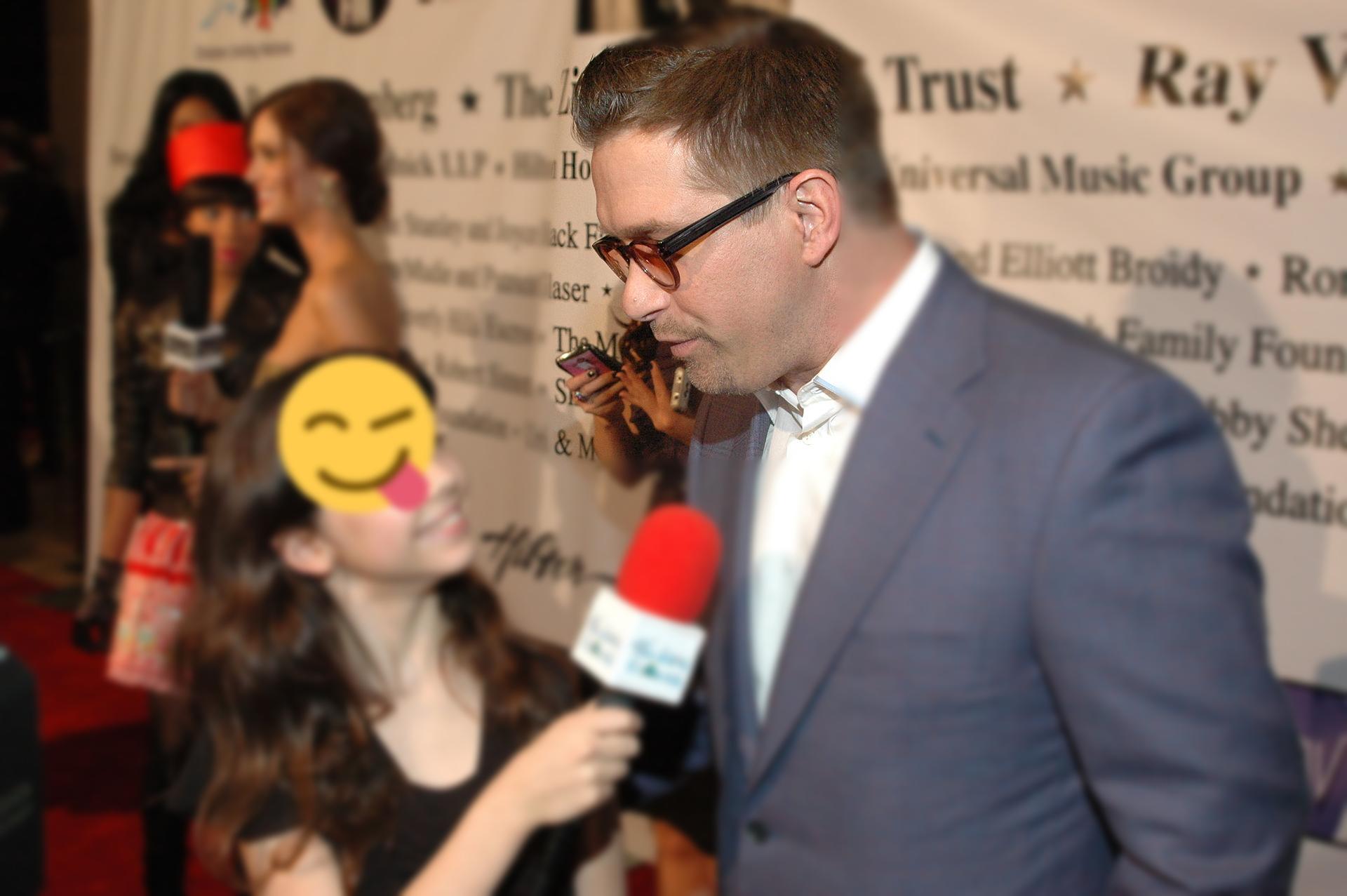}
\hfill
\begin{lstlisting}[frame=single]
Instruction: Hide Jennifer Smart 
with face_savoring_food and blur 
the background of Stephen 
Baldwin.
\end{lstlisting}

\end{subfigure}
\caption{Image Editing (OFCP) exemplars.}
\label{fig:img_editing_ofcp}
\end{figure*}

\begin{figure*}
    \centering
    
    \begin{subfigure}{0.48\linewidth}
    \begin{subfigure}{0.33\linewidth}
    \centering
    \includegraphics[width=\textwidth]{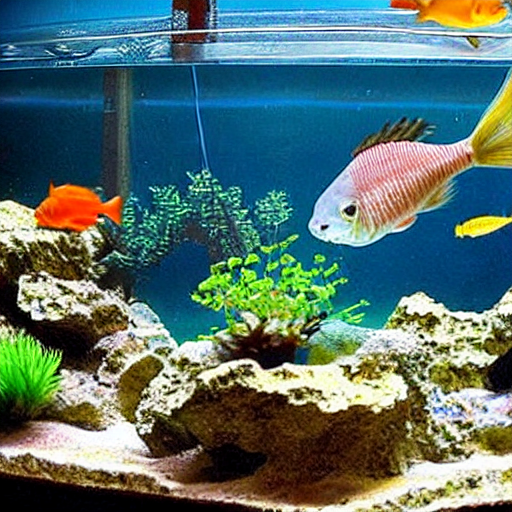}
    \end{subfigure}\hfill
    \begin{subfigure}{0.33\linewidth}
    \centering
    \includegraphics[width=\textwidth]{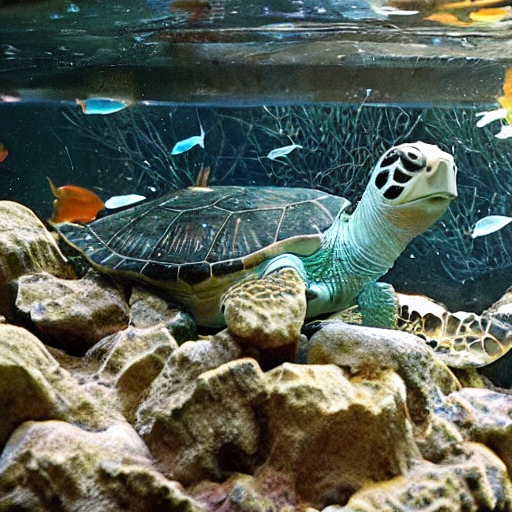}
    \end{subfigure}\hfill
    \begin{subfigure}{0.33\linewidth}
    \centering
    \includegraphics[width=\textwidth]{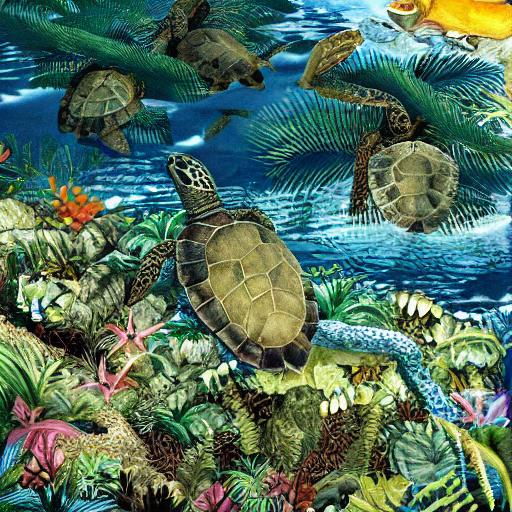}
    \end{subfigure}
    \begin{lstlisting}[frame=single,xrightmargin=5px]
Instruction: 
Start with an image of a fish in an aquarium. 
Then replace the fish with a turtle. 
Then refine the aquarium with tropical setup. 
    \end{lstlisting}
    \begin{lstlisting}[frame=single,xrightmargin=5px]
Programmatic Query:
RefineObject(
   ReplaceObject(
       Background("a fish in an aquarium"), "fish", "turtle"), 
       "aquarium",  "tropical setup")
    \end{lstlisting}
    \end{subfigure}
    \hfill
    \begin{subfigure}{0.48\linewidth}
    \begin{subfigure}{0.33\linewidth}
    \centering
    \includegraphics[width=\textwidth]{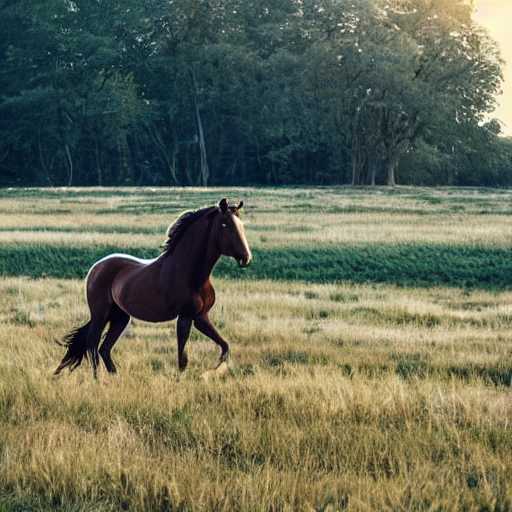}
    \end{subfigure}\hfill
    \begin{subfigure}{0.33\linewidth}
    \centering
    \includegraphics[width=\textwidth]{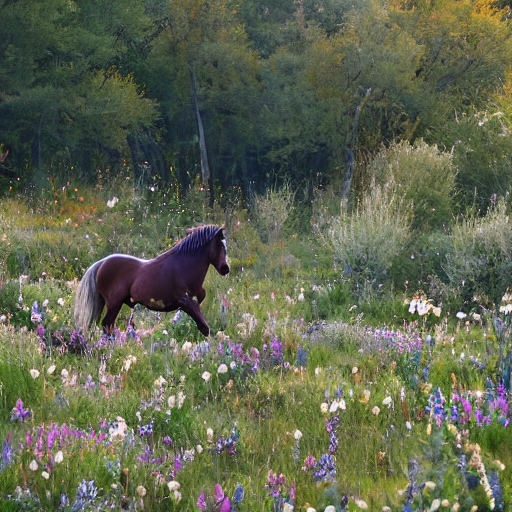}
    \end{subfigure}\hfill
    \begin{subfigure}{0.33\linewidth}
    \centering
    \includegraphics[width=\textwidth]{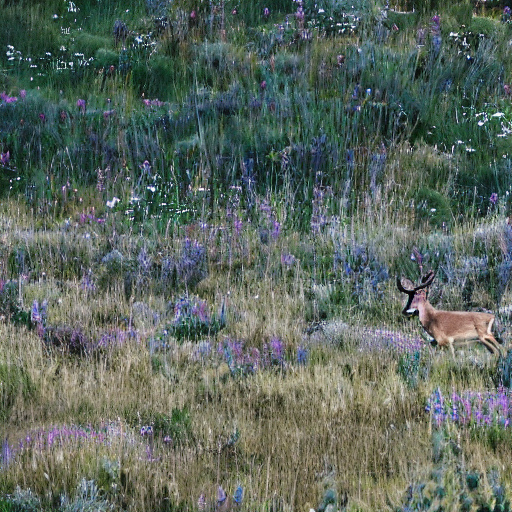}
    \end{subfigure}
    \begin{lstlisting}[frame=single,xrightmargin=5px]
Instruction: 
Start with an image of a horse in a meadow. 
Refine the meadow to have wildflowers and 
replace the horse with a deer. 
    \end{lstlisting}
    \begin{lstlisting}[frame=single,xrightmargin=5px]
Programmatic Query:
ReplaceObject(
    RefineObject(
        Background("a horse in a meadow"), 
        "meadow", "meadow with wildflowers"), 
    "horse", "deer")
    \end{lstlisting}
    \end{subfigure}

    \begin{subfigure}{0.48\linewidth}
    \begin{subfigure}{0.33\linewidth}
    \centering
    \includegraphics[width=\textwidth]{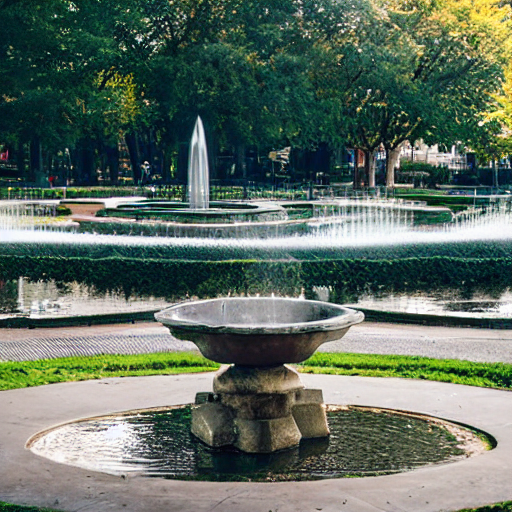}
    \end{subfigure}\hfill
    \begin{subfigure}{0.33\linewidth}
    \centering
    \includegraphics[width=\textwidth]{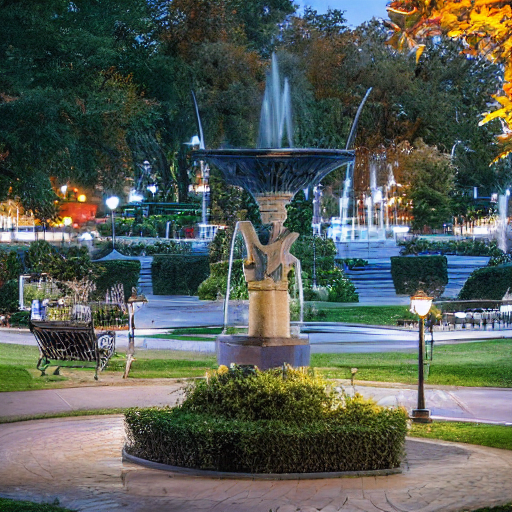}
    \end{subfigure}\hfill
    \begin{subfigure}{0.33\linewidth}
    \centering
    \includegraphics[width=\textwidth]{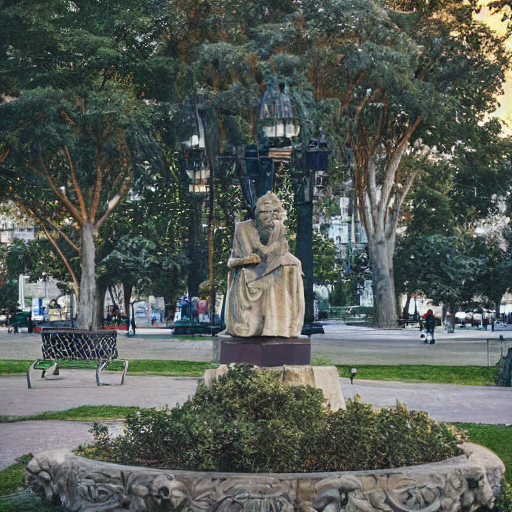}
    \end{subfigure}
    \begin{lstlisting}[frame=single,xrightmargin=5px]
Instruction: 
Start with an image of a park with a fountain. 
Replace the fountain with a statue and refine
the park to evening.
    \end{lstlisting}
    \begin{lstlisting}[frame=single,xrightmargin=5px]
Programmatic Query:
ReplaceObject(
    RefineObject(
        Background("a park with a fountain"), "park", "park in the evening"), 
        "fountain", "statue")
    \end{lstlisting}
    \end{subfigure}
    \hfill
    \begin{subfigure}{0.48\linewidth}
    \begin{subfigure}{0.33\linewidth}
    \centering
    \includegraphics[width=\textwidth]{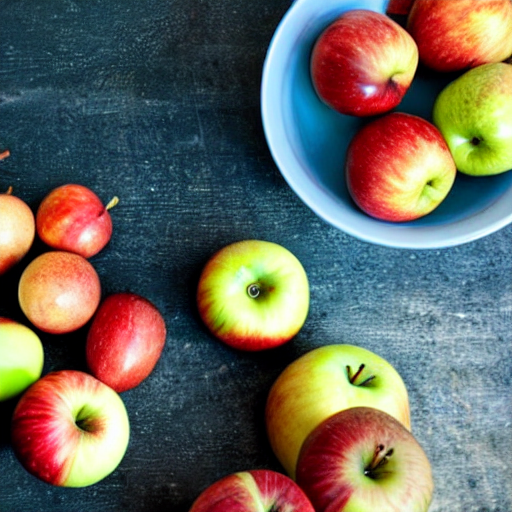}
    \end{subfigure}\hfill
    \begin{subfigure}{0.33\linewidth}
    \centering
    \includegraphics[width=\textwidth]{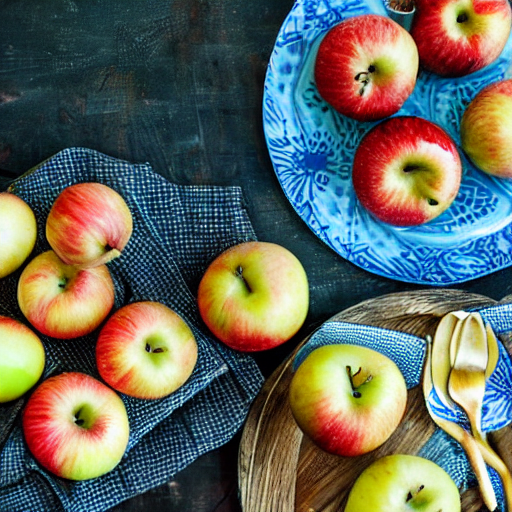}
    \end{subfigure}\hfill
    \begin{subfigure}{0.33\linewidth}
    \centering
    \includegraphics[width=\textwidth]{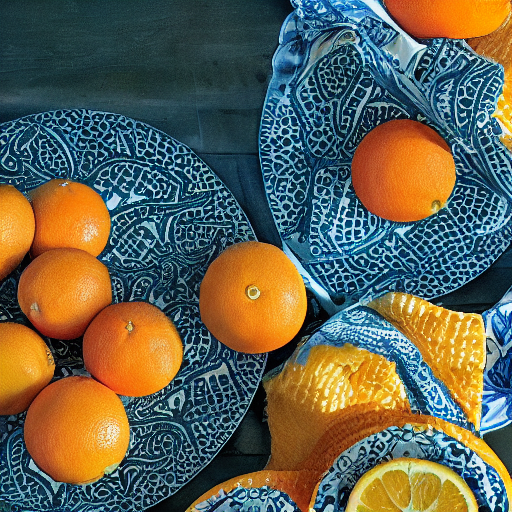}
    \end{subfigure}
    \begin{lstlisting}[frame=single,xrightmargin=5px]
Instruction: 
Start with an image of a bowl full of apples. 
Then replace the bowl with something else, and 
change the apples to other fruits.
    \end{lstlisting}
    \begin{lstlisting}[frame=single,xrightmargin=5px]
Programmatic Query:
NotObject(
    NotObject(
        Background("a bowl full of apples"),
        "bowl"), 
    "apples")
    \end{lstlisting}
    \end{subfigure}

    \begin{subfigure}{0.48\linewidth}
    \begin{subfigure}{0.33\linewidth}
    \centering
    \includegraphics[width=\textwidth]{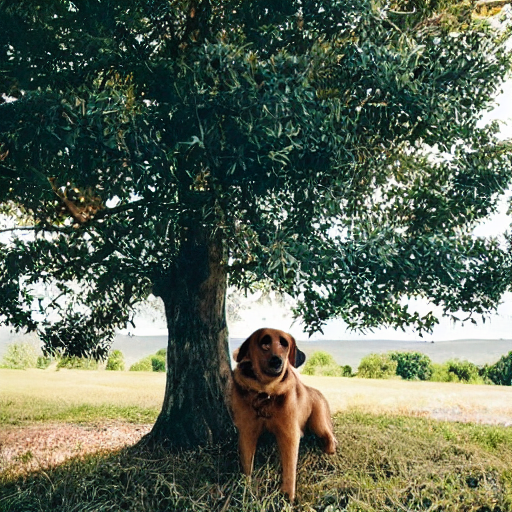}
    \end{subfigure}\hfill
    \begin{subfigure}{0.33\linewidth}
    \centering
    \includegraphics[width=\textwidth]{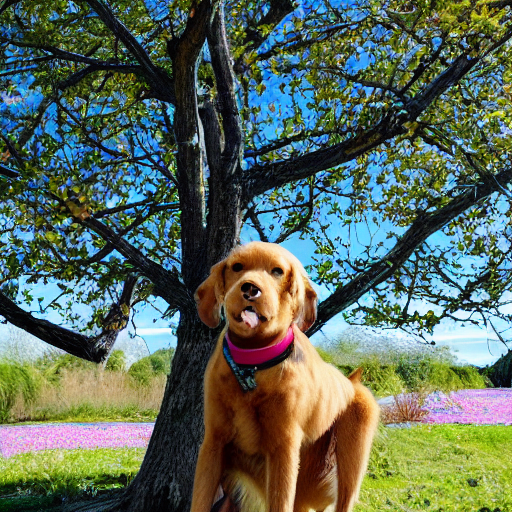}
    \end{subfigure}\hfill
    \begin{subfigure}{0.33\linewidth}
    \centering
    \includegraphics[width=\textwidth]{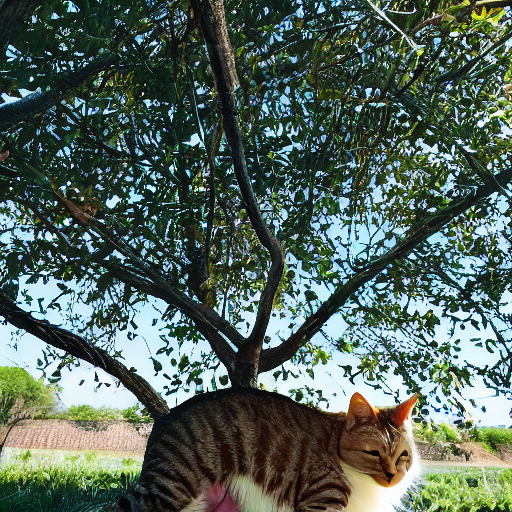}
    \end{subfigure}
    \begin{lstlisting}[frame=single,xrightmargin=5px]
Instruction: 
Start with an image of a dog under a tree. Then 
refine the season to spring, and replace the 
dog to something else.
    \end{lstlisting}
    \begin{lstlisting}[frame=single,xrightmargin=5px]
Programmatic Query:
NotObject(
    RefineObject(
        Background("a dog under a tree"), 
        "tree", "tree in spring"), 
    "dog")
    \end{lstlisting}
    \end{subfigure}
    \hfill
    \begin{subfigure}{0.48\linewidth}
    \begin{subfigure}{0.33\linewidth}
    \centering
    \includegraphics[width=\textwidth]{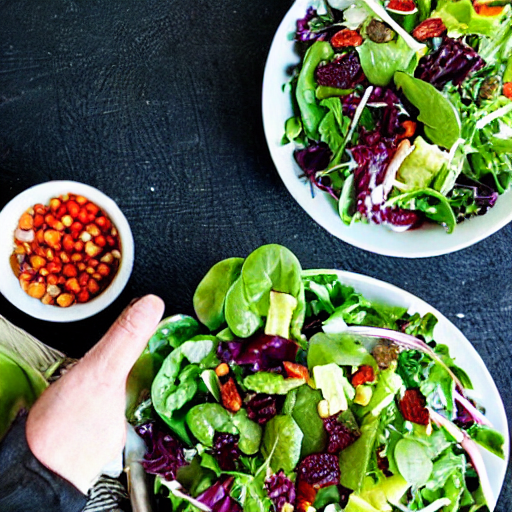}
    \end{subfigure}\hfill
    \begin{subfigure}{0.33\linewidth}
    \centering
    \includegraphics[width=\textwidth]{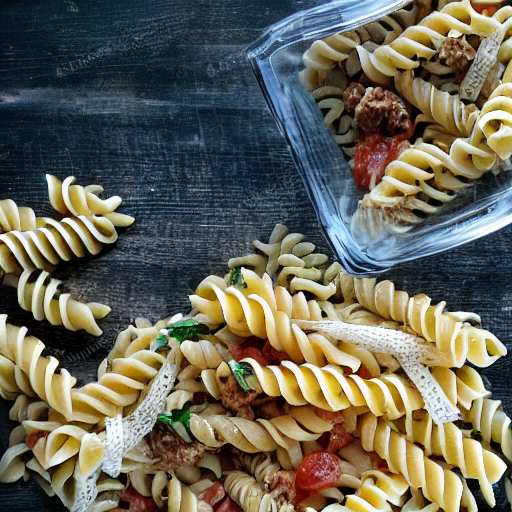}
    \end{subfigure}\hfill
    \begin{subfigure}{0.33\linewidth}
    \centering
    \includegraphics[width=\textwidth]{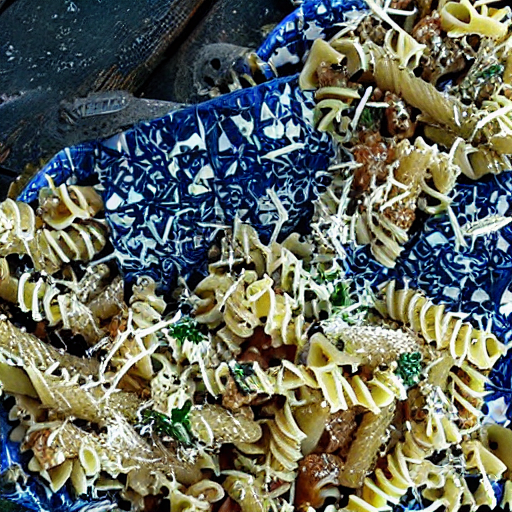}
    \end{subfigure}
    \begin{lstlisting}[frame=single,xrightmargin=5px]
Instruction: 
Start with an image of a bowl of salad. Replace 
the salad with pasta and replace the bowl to
something else. 
    \end{lstlisting}
    \begin{lstlisting}[frame=single,xrightmargin=5px]
Programmatic Query:
RefineObject(
    ReplaceObject(
        Background("a bowl of salad"), 
        "salad", "pasta"), 
    "bowl", "plate")
    \end{lstlisting}
    \end{subfigure}
    \caption{Image Generation and Editing (IGP20) exemplars.}
    \label{fig:img-gen-qualitative}
\end{figure*}

We present exemplars for face tagging in Figure \ref{fig:face_tagging}, object tagging in Figure \ref{fig:object_tagging}, image editing in Figure \ref{fig:img_editing_ofcp}, and image generation and editing in Figure \ref{fig:img-gen-qualitative}.

\end{document}